\pdfoutput=1

\documentclass[11pt]{article}

\usepackage[final]{acl}

\usepackage{times}
\usepackage{latexsym}

\usepackage[T1]{fontenc}

\usepackage[utf8]{inputenc}

\usepackage{microtype}

\usepackage{inconsolata}

\usepackage{graphicx}

\usepackage{booktabs}
\usepackage{amsfonts,amssymb}
\usepackage{amsmath,amsthm}
\usepackage{mathrsfs}
\usepackage{latexsym}
\usepackage{multirow} 
\usepackage{multicol}
\usepackage{arydshln}
\usepackage{colortbl}
\usepackage{color}
\usepackage{xcolor}
\usepackage{array}
\usepackage{caption}
\usepackage{makecell}

\title{Mitigating Biases of Large Language Models in Stance Detection with Counterfactual Augmented Calibration}

\author{Ang Li$^{1,3\ast}$, Jingqian Zhao$^{1}$\thanks{\quad The first two authors contribute equally to this work.}, Bin Liang$^{1,3,4}$, Lin Gui$^{5}$, Hui Wang$^{2}$, \\ \bf Xi Zeng$^{6}$, Xingwei Liang$^{1}$, Kam-Fai Wong$^{3,4}$, and Ruifeng Xu$^{1,2}$\thanks{\quad Corresponding Author}\\
    $^{1}$ Harbin Institute of Technology, Shenzhen, China~
    $^{2}$ Peng Cheng Laboratory, Shenzhen, China \\
    $^{3}$ The Chinese University of Hong Kong, Hong Kong, China \\
    $^{4}$ MoE Key Lab of High Confidence Software Technologies, CUHK, China \\
    $^{5}$ King’s College London, UK~
    $^{6}$ The 30th Research Institute of \\ China Electronics Technology Group Corporation, Chengdu, China \\
    \texttt{\{angli,zhaojingqian\}@stu.hit.edu.cn},
    ~\texttt{xuruifeng@hit.edu.cn} \\
}

\begin{document}
\maketitle
\begin{abstract}
Stance detection is critical for understanding the underlying position or attitude expressed toward a topic.
Large language models (LLMs) have demonstrated significant advancements across various natural language processing tasks including stance detection, however, their performance in stance detection is limited by biases and spurious correlations inherent due to their data-driven nature.
Our statistical experiment reveals that LLMs are prone to generate biased stances due to sentiment-stance spurious correlations and preference towards certain individuals and topics. Furthermore, the results demonstrate a strong negative correlation between stance bias and stance detection performance, underscoring the importance of mitigating bias to enhance the utility of LLMs in stance detection.
Therefore, in this paper, we propose a Counter\textbf{fact}ual A\textbf{u}gmented C\textbf{al}ibration Network (\texttt{FACTUAL}), which a novel calibration network is devised to calibrate potential bias in the stance prediction of LLMs. Further, to address the challenge of effectively learning bias representations and the difficulty in the generalizability of debiasing, we construct counterfactual augmented data. This approach enhances the calibration network, facilitating the debiasing and out-of-domain generalization. Experimental results on in-target and zero-shot stance detection tasks show that the proposed \texttt{FACTUAL} can effectively mitigate biases of LLMs, achieving state-of-the-art results.
\end{abstract}

\section{Introduction}

Stance detection aims at automatically identifying the author's opinionated standpoint or attitude (e.g., {\em Favor}, {\em Against}, or {\em Neutral}) expressed in the content towards a specific target, topic, or proposition~\cite{somasundaran-wiebe-2010-recognizing, DBLP:conf/semeval/MohammadKSZC16}. With the development of social media platforms, stance detection plays a pivotal role in analyzing public opinion on social media topics~\cite{DBLP:conf/sigir/JangA18, DBLP:conf/clef/GhoshSSRG19, DBLP:conf/acl/StefanovDAN20,DBLP:conf/coling/SunWZZ18, DBLP:conf/icann/ChenYC21}. 

Large Language Models (LLMs), such as ChatGPT\footnote{\url{https://openai.com/blog/chatgpt/}}, Bard\footnote{\url{https://bard.google.com/}}, and LLaMA~\cite{DBLP:journals/corr/abs-2302-13971}, have demonstrated impressive language comprehension and task-handling capabilities by leveraging extensive corpus and knowledge. 
However, their data-driven nature makes them susceptible to biases and spurious correlations embedded in pre-training data. In stance detection, which requires interpreting the relationship between a sentence and a specific topic, clues from any isolated aspects could become spurious and lead to biased stances.

\begin{figure}[!t]
\centering  
\includegraphics[width=1\linewidth]{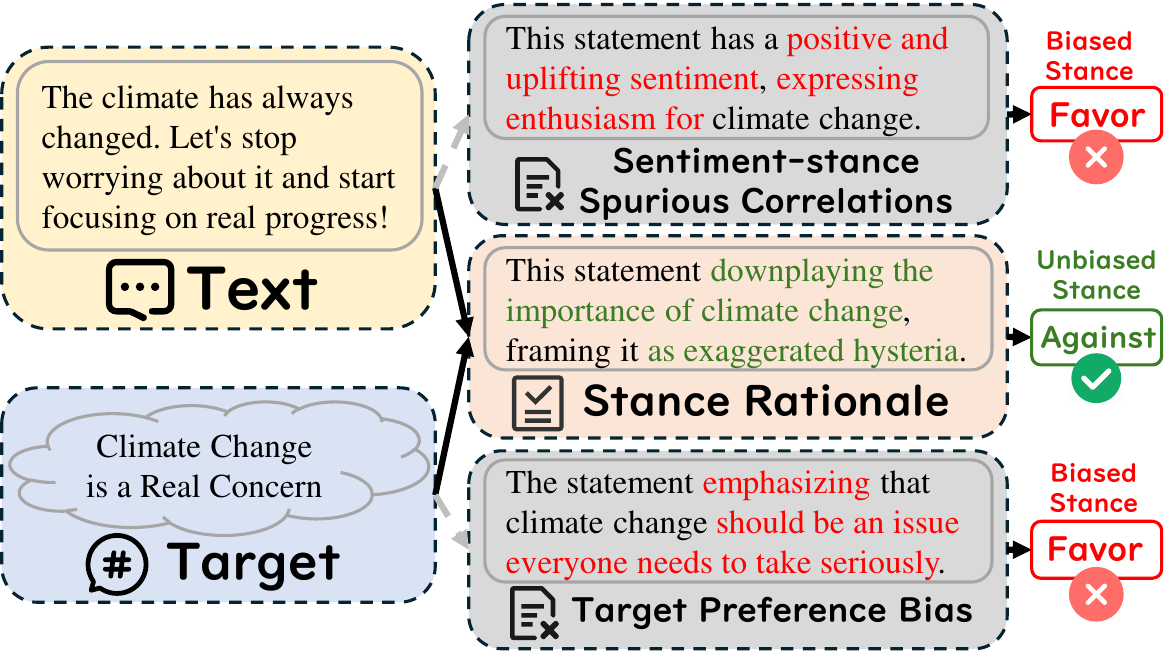}
\caption{An example demonstrates two types of biases encountered by large language models in stance detection tasks (shown at the top and bottom) as well as unbiased stance rationale (shown in the middle).} 
\label{Fig:intro_example}
\end{figure}

Our experiment identifies two primary biases in LLMs for stance detection: (1) sentiment-stance spurious correlations, where sentiment misleads stance judgment, and (2) target preference bias, where LLMs favor certain individuals or topics. Figure~\ref{Fig:intro_example} illustrates examples of the two types of biases, as well as unbiased stance rationale.
Furthermore, our results reveal a significant negative correlation between stance bias and stance detection performance, emphasizing the necessity of alleviating bias to improve the effectiveness of LLMs in stance detection.

Existing research of debiasing in stance detection largely centered on the creation of unbiased training samples and the retraining of stance detection models~\cite{DBLP:conf/naacl/KaushalSG21, DBLP:journals/corr/abs-2212-10392}. However, there are two core limitations to the application of these debiasing methods in LLMs. \textbf{Limitation\#1}, research~\cite{DBLP:journals/corr/abs-2308-08747} has shown that such retraining processes will undermine the generality of LLMs, potentially leading to catastrophic forgetting; not to mention that there are restrictions with certain closed-source LLMs like GPT-3.5-turbo, which can only be accessed with a restricted inference API, preventing access to internal model parameters. \textbf{Limitation\#2}, existing approaches to constructing unbiased training samples typically entail the analysis of prevalent bias patterns, subsequently automating their construction based on these identified patterns, exemplified by substituting `Men' with `Women'. However, when dealing with stance detection tasks, our forthcoming analysis illuminates that these samples display varying bias propensities, attributable to divergences in sentiments and stance objectives. Consequently, utilizing conventional methods to create unbiased samples poses a significant challenge.

Therefore, to address the above two limitations, we propose to mitigate biases of LLMs in stance detection with a Counter\textbf{fact}ual A\textbf{u}gmented C\textbf{al}ibration Network, coined as \texttt{FACTUAL}.
We establish a trainable calibration network to approximate the inverse projection function of the bias label distribution within LLMs. This calibration network takes samples as input, including stance judgments and rationales from LLMs, and generates calibrated stance judgments.
We construct counterfactual augmented data against the training data to rectify stance biases. The counterfactual samples are constructed from both causal and non-causal features, which can enhance the calibration network to yield unbiased stances and accomplish out-of-domain generalization.
Through counterfactual augmented supervised training, the calibration network can capture biases present in specific samples, thereby performing debiasing. The main contributions of our work are summarized as follows:

1) We are the first to investigate the biases of LLMs on stance detection, categorizing the biases into two main types from the perspective of causality and proposing metrics to quantify these two types of biases.

2) We propose \texttt{FACTUAL}, a novel framework called the Counterfactual Augmented Calibration Network to mitigate biases of LLMs on stance detection.

3) A series of experiments demonstrate that our \texttt{FACTUAL} can effectively reduce the bias of LLMs in stance detection, improving the performance in both in-target and zero-shot stance detection tasks\footnote{The code is available at \url{https://github.com/Leon-Francis/FACTUAL}.}.

\section{Related Work}

\paragraph{Biases in Large Language Models}
Some studies~\cite{DBLP:conf/emnlp/GoncalvesS23} have examined the biases existing in Large Language Models (LLMs), these biases mainly include gender and religion~\cite{DBLP:journals/corr/abs-2310-08780}, politics~\cite{DBLP:journals/corr/abs-2311-08605, DBLP:journals/corr/abs-2311-09687}, and spurious correlations~\cite{DBLP:journals/corr/abs-2311-08648}. The associated debiasing efforts are centered around retraining the language model with debiased samples~\cite{DBLP:journals/corr/abs-2310-12490, DBLP:journals/corr/abs-2310-18913}.
\citet{DBLP:journals/corr/abs-2309-03882} found that LLMs are vulnerable to option position changes in MCQs due to their inherent `selection bias'. They perform debiasing by approximating the overall bias distribution. While based on our analysis in Section~\ref{bias-llms}, the bias distribution varies significantly across different stance detection samples, so this method is not applicable.

\paragraph{Mitigating Biases in Stance Detection}
Currently, studies developed for mitigating biases in stance detection are oriented toward fine-tuned models. \citet{DBLP:conf/naacl/KaushalSG21} analyzed two biases existing in the current datasets: target-independent lexical choices and target-independent sentiment-stance correlations, and built an unbiased dataset. \citet{DBLP:conf/coling/YuanZL022} incorporated the stance reasoning process as task knowledge to retrain the model to reduce bias. \citet{DBLP:journals/corr/abs-2212-10392} constructed unbiased samples through counterfactual reasoning and performed adversarial bias learning. These methods involve retraining models and constructing unbiased training samples through special marks, which cannot be directly applied to LLMs.

\section{Biases of LLMs in Stance Detection}
\label{bias-llms}

\subsection{Bias Measurement}
Stance bias refers to the systematic errors where models tend to choose certain stances due to the influence of specific biases and stereotypes. Inspired by~\citet{DBLP:journals/corr/abs-2309-03882}, the standard deviation of recalls (\textit{RStd}) on stance labels is an excellent metric for quantitatively measuring systematic errors. The formula is as follows:
\begin{equation}
    RStd = \sqrt{\frac{1}{K} \sum_{i=1}^{K} \left(\frac{TP_i}{P_i} - \frac{1}{K} \sum_{j=1}^{K} \frac{TP_j}{P_j}\right)^2}
\end{equation}
Where $K$ is the number of stance labels, $TP_i$ is the number of true positive instances for stance label $i$, and $P_i$ is the number of instances of stance label $i$. This measurement resists label imbalance and effectively reflects the model's bias tendency on samples (Refer to Appendix~\ref{sec:measurement_analysis} for the validation).

\subsection{Experimental Result}
\label{subsec:exp_result}

Through statistical analysis of the results from LLMs, we identified two significant types of bias: \textbf{Sentiment-stance Spurious Correlations} and \textbf{Target Preference Bias}.

\subsubsection{Sentiment-Stance Spurious Correlations}
\label{subsubsec: SSC}

\begin{figure}[!t]
\centering  
\includegraphics[width=1\linewidth]{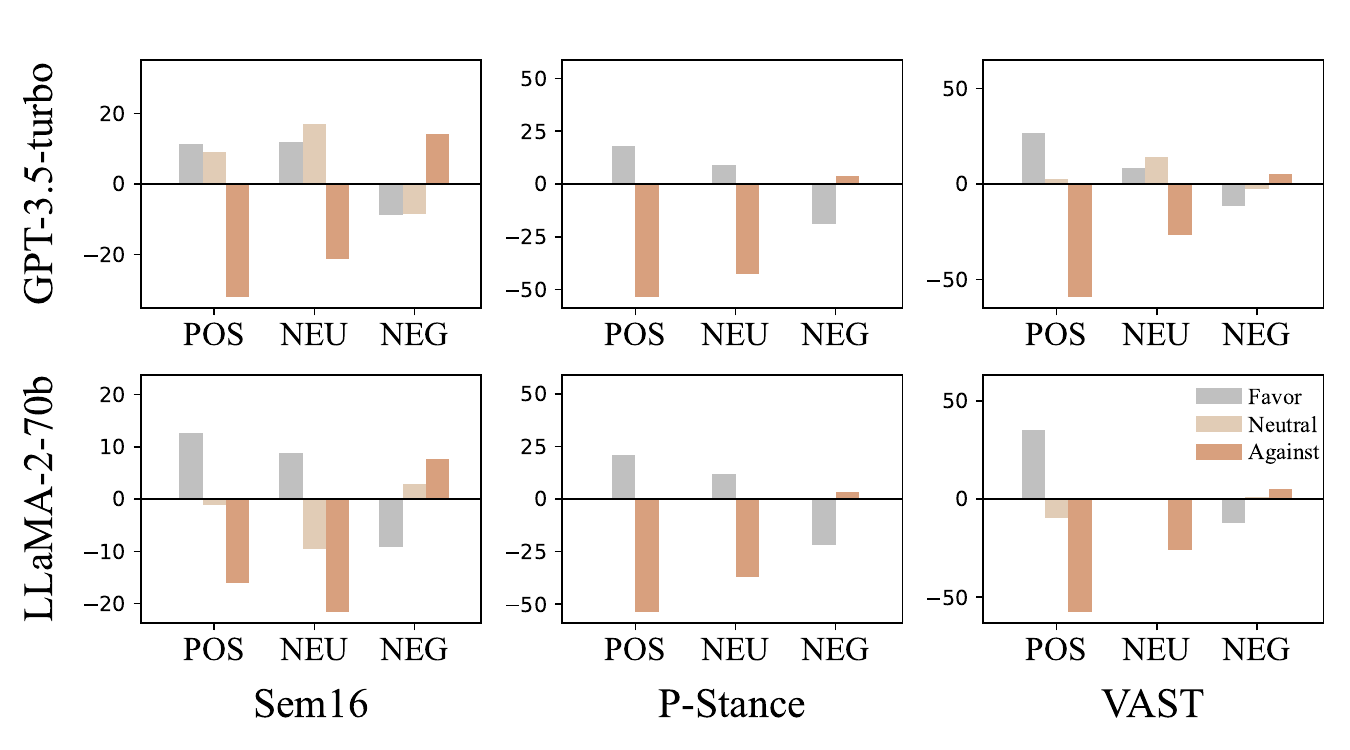}
\caption{The recall score of each stance label on three sentiment subsets, normalizing by subtracting the overall recall scores of the corresponding stance labels across overall dataset, on Sem16, P-Stance, and VAST. POS for positive, NEU for neutral, NEG for negative.}
\label{Fig:sent-spurious}
\end{figure}

Sentiment can influence stance judgment but is not the primary determinant. Overreliance on sentiment by the model suggests susceptibility to sentiment-stance spurious correlations, leading to biased stance assessments.
To investigate stance bias across different sentiments, we first ascertain the sentiment label for each sample. In the Sem16 dataset, each sample has annotated sentiment labels, categorized as positive, neutral, or negative. For the P-Stance and VAST datasets, we utilize GPT-4 to annotate the sentiment labels. To gain a preliminary understanding of sentiment-stance spurious correlations, we first divide the dataset into three subsets based on sentiment categories. For each subset, we calculate the recall score for each stance label and normalize it by subtracting the overall recall score of the corresponding stance labels across the dataset, as shown in Figure~\ref{Fig:sent-spurious}. Observations indicate that LLMs tend to erroneously predict `support' for positively-sentiment samples and `against' for negatively-sentiment ones, indicating a deviation from expected patterns and highlighting an inherent stance bias.
Hence, we identify the \textbf{S}entiment-stance \textbf{S}purious \textbf{C}orrelations (SSC) as a type of bias in LLMs on stance detection.

We calculate the average of the RStd across all sentiments as our quantification for sentiment-stance spurious correlations:
\begin{equation}
    \textit{Bias-SSC} = \frac{1}{|S|} \sum_{s \in S} RStd(X_s)
\end{equation}
where $X_s$ represents instances with sentiment label $s$, $|S|$ denotes the number of sentiment labels, which in our experiment, is 3. 

\begin{figure}[!t]
\centering  
\includegraphics[width=1\linewidth]{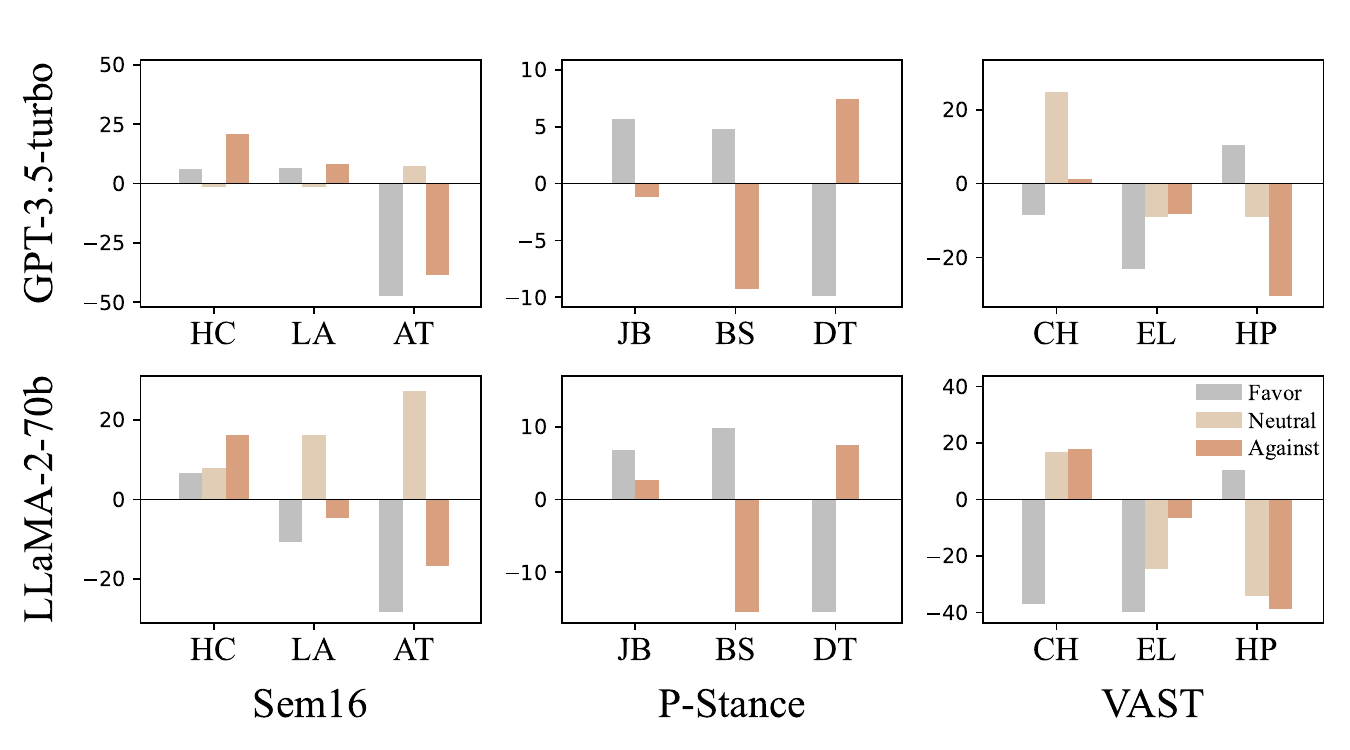}
\caption{The recall score of each stance label on several target subsets, normalizing by subtracting the overall recall score of the corresponding stance labels across all targets, on Sem16, P-Stance, and VAST dataset. HC for Hillary Clinton, LA for Legalization of Abortion, AT for Atheism, JB for Joe Biden, BS for Bernie Sanders, DT for Donald Trump, CH for Christian, CL for Election, HP for Humanity Program.}
\label{Fig:target-prefer}
\end{figure}

We conducted experiments in various settings: \textit{Task-Des} used task-related descriptions for stance judgment\footnote{The results presented in Figures~\ref{Fig:sent-spurious} and~\ref{Fig:target-prefer} are obtained based on \textit{Task-Des}.}, \textit{CoT-Demo} used the task description with 4-shot chain-of-thought demonstration, and \textit{Debias-Instruct} used the task description indicating that sentiment was spurious cues for stance judgment. Refer to Appendix~\ref{sec:prompt_setting} for the detailed prompts. The results are shown in Table~\ref{Tab:sent-spurious-result}. We can observe that in most cases, there is a negative correlation between bias-SSC and stance detection performance. See further analysis in Appendix~\ref{sec:bias_influence}. Moreover, prompt engineering methods proved ineffectual in mitigating this inherent bias.

\begin{table}[!t]
\small
\centering
\setlength{\tabcolsep}{2.2pt}
\renewcommand{\arraystretch}{1}
\begin{tabular}{lcccccccc}
\hline
                   & \multicolumn{2}{c}{Sem16} &  & \multicolumn{2}{c}{P-Stance} &  & \multicolumn{2}{c}{VAST} \\
                   \cline{2-3} \cline{5-6} \cline{8-9}
                   & SSC$\downarrow$      & F1$\uparrow$       &  & SSC$\downarrow$      & F1$\uparrow$         &  & SSC$\downarrow$    & F1$\uparrow$      \\
\hline
\multicolumn{9}{l}{\textbf{LLaMA-2-70b-chat}}                                                                           \\
Task-Des           & \textbf{17.80}         & 60.08     &  & 23.36          & 79.89       &  & \textbf{16.87}        & 68.36     \\
CoT-Demo    & 27.52         & 58.68     &  & \textbf{22.81}          & \textbf{80.77}       &  & 22.55        & 67.08     \\
Debias-Instruct & 19.24         & \textbf{63.62}     &  & 24.86          & 78.85       &  & 19.63        & \textbf{68.68}     \\
\hline
\multicolumn{9}{l}{\textbf{GPT-3.5-Turbo-0125}}                                                                         \\
Task-Des           & 27.13         & 52.82     &  & 23.72          & \textbf{81.62}       &  & 28.70        & 49.86     \\
CoT-Demo    & \textbf{18.08}         & \textbf{67.59}     &  & \textbf{22.75}          & 80.88       &  & \textbf{16.32}        & \textbf{69.90}     \\
Debias-Instruct & 23.75         & 51.77     &  & 23.48          & 81.48       &  & 30.53        & 48.68     \\
\hline
\end{tabular}
\caption{Bias-SSC and macro F1-score of stance detection on the Sem16, P-Stance and VAST dataset. Refer to Appendix~\ref{sec:experimental_result} for detailed results on each sentiment.}
\label{Tab:sent-spurious-result}
\end{table}

\subsubsection{Target Preference Bias}
\label{subsubsec: TPB}

LLMs exhibit bias towards certain individuals or topics. This bias can interfere with their ability to judge stances based on the text, leading to biased stance judgments. We refer to this bias as target preference bias. To preliminarily observe the target preference bias of LLMs, we randomly sampled some targets from different datasets, calculated the recall scores for each stance label on each target subset, and normalized it by subtracting the overall recall score of the corresponding stance labels across all targets, as shown in Figure~\ref{Fig:target-prefer}.
We observed that, on different targets, LLMs displayed markedly different tendencies in stance selection, which ultimately affected the correctness of stance judgment. Therefore, we identify the \textbf{T}arget \textbf{P}reference \textbf{B}ias (TPB) as a type of bias in LLMs on stance detection.

We calculate the average of the RStd of all targets as our quantification for target preference bias:
\begin{equation}
    \textit{Bias-TPB} = \frac{1}{|T|} \sum_{t \in T} RStd(X_t)
\end{equation}
where $X_t$ represents instances with stance target $t$, $|T|$ denotes the number of targets.

We conduct experiments based on \textit{Task-Des}, \textit{CoT-Demo}, and \textit{Debias-Instruct} which emphasize the need to judge the stance based on the text and not to include the inherent attitude towards the target. Refer to Appendix~\ref{sec:prompt_setting} for the detailed prompts. The results are shown in Table~\ref{Tab:target-prefer-result}. We can observe that in most cases, there is a negative correlation between bias-TPB and stance detection performance. See further analysis in Appendix~\ref{sec:bias_influence}, and prompt engineering fails to effectively mitigate bias-TPB.

\begin{table}[!t]
\small
\centering
\setlength{\tabcolsep}{2.2pt}
\renewcommand{\arraystretch}{1}
\begin{tabular}{lcccccccc}
\hline
                & \multicolumn{2}{c}{Sem16}       &  & \multicolumn{2}{c}{P-Stance}   &  & \multicolumn{2}{c}{VAST}       \\
                \cline{2-3} \cline{5-6} \cline{8-9}
                & TPB$\downarrow$            & F1$\uparrow$             &  & TPB$\downarrow$           & F1$\uparrow$             &  & TPB$\downarrow$           & F1$\uparrow$             \\
\hline
\multicolumn{9}{l}{\textbf{LLaMA-2-70b-chat}}                                                                                      \\
Task-Des        & 17.59          & 60.08          &  & 9.09          & 79.89          &  & 7.76          & 68.36          \\
CoT-Demo        & 27.56          & 58.68          &  & 11.57         & \textbf{80.77} &  & 9.64          & 67.08          \\
Debias-Instruct & \textbf{16.37} & \textbf{61.40} &  & \textbf{8.94} & 78.70          &  & \textbf{4.86} & \textbf{69.10} \\
\hline
\multicolumn{9}{l}{\textbf{GPT-3.5-Turbo-0125}}                                                                                    \\
Task-Des        & 22.64          & 52.82          &  & \textbf{5.43} & \textbf{81.62} &  & 28.44         & 49.86          \\
CoT-Demo        & \textbf{13.47} & \textbf{67.59} &  & 6.61          & 80.88          &  & \textbf{8.40} & \textbf{69.90} \\
Debias-Instruct & 21.87          & 53.33          &  & 5.79          & 81.59          &  & 26.77         & 51.66          \\
\hline
\end{tabular}
\caption{Bias-TPB and macro F1-score of stance detection on the Sem16, P-Stance and VAST dataset. Refer to Appendix~\ref{sec:experimental_result} for detailed results on each target.}
\label{Tab:target-prefer-result}
\end{table}

\begin{figure*}[!t]
\centering  
\includegraphics[width=1\linewidth]{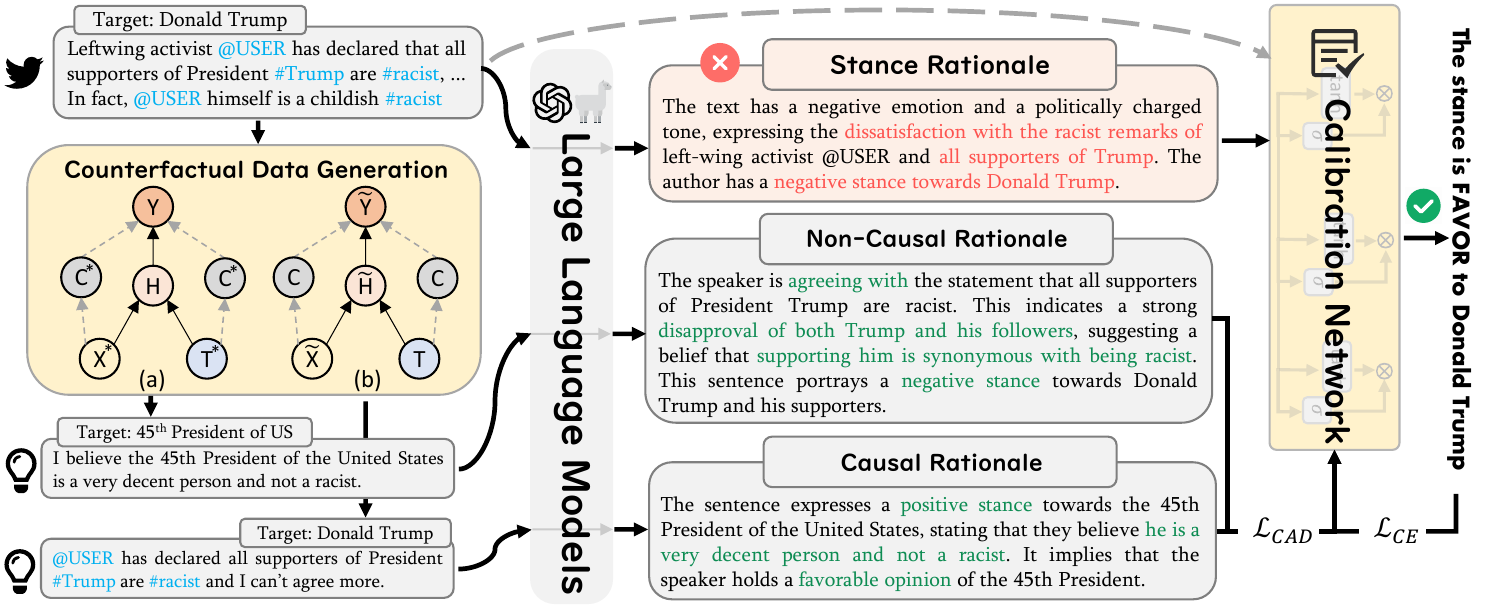}
\caption{The overall architecture of our proposed \texttt{FACTUAL}. (a) and (b) in the counterfactual data generation represent two ways to generate counterfactual augmentation. $X$ donates the text, $T$ donates the target, $H$ donates the features of the interaction of text and target, and $Y$ donates the stance label. $C$ represents confounding factors, which arise from the two types of biases previously analyzed and may distort the stance prediction. $*$ denotes the perturbation of non-causal features, and $\sim$ denotes the perturbation of causal features.} 
\label{Fig:framework}
\end{figure*}

\section{Mitigating Bias with Calibration}

Given $\{x_n, t_n\}_{n=1}^{N}$ as the labeled dataset, where $x$ denotes the input text and $t$ denotes the corresponding target, LLMs obtain the stance predictions $y$ through the task instructions $\mathcal{I}$ for stance detection: $P_{obs}(y_{i}|\mathcal{I};x_{i}, t_{i})$. Inspired by~\citet{DBLP:journals/corr/abs-2309-03882}, we believe that it can be deconstructed into the unbiased distribution $P_{unbiased}$ of the LLMs performing the stance detection task, and the bias distribution $P_{bias}$ formed by confounding $C_i$:
\begin{equation}
    P_{obs} = P_{unbiased}(y_{i}|x_{i}, t_{i})P_{bias}(y_{i}|C_i)
\end{equation}
The confounding factor $C_i$ arises from the two types of biases analyzed earlier. It is important to note that it may be unaffected, influenced by a single bias, or impacted by both types of biases.
Specifically, $P_{obs}$ denotes the stance judgment text, which can be regarded as the outcome of word probability distribution following the argmax operation, derived from LLMs. We aim to estimate the unbiased stance distribution $P_{unbiased}$.

\subsection{Calibration Network}

By estimating the bias distribution based on the overall distribution of known samples (from the training set), we can obtain unbiased outputs by multiplying the observed distribution of LLMs by the inverse of the approximated bias distribution:
\begin{equation}
\scalebox{0.95}{
$\displaystyle
    P_{unbiased} = P_{obs}(y_i^{\prime}|\mathcal{I}; x_i, t_i) \widetilde{P}_{bias}(y_i^{\prime \prime}|C_i)^{-1}
$}
\end{equation}
where $y_i^{\prime}$ and $y_i^{\prime \prime}$ represent the label distribution output by LLMs and the label distribution affected by bias, $\widetilde{P}_{bias}$ represents the estimate of bias.

However, based on the bias analysis in Section~\ref{bias-llms}, we found that for stance detection, the samples with different sentiments and stance targets have completely different stance bias distributions. Therefore, we propose employing a network to capture the bias distribution specific to each sample, called the \textit{calibration network} $f_{Cal}$. We use the network $f_{Cal}$ to approximate the inverse projection function of the bias distribution:
\begin{equation}
     f_{Cal} = P_{bias}(y_i^{\prime \prime}|C_i)^{-1}
\end{equation}
By inputting the predicted stance distribution $P_{obs}$ from LLMs, an approximating unbiased label can be obtained:
\begin{equation}
     P_{unbiased}(\hat{y_i}) = f_{Cal}(P_{obs}(y_i^{\prime}|\mathcal{I}; x_i, t_i))
\end{equation}
Specifically, as illustrated in Figure~\ref{Fig:framework}, we first use the \textit{CoT-Demo} instruction (refer to Appendix~\ref{sec:prompt_setting} for the detail) to obtain the stance judgment and rationale from the LLMs (which correspond to the result of an argmax operation over the $P_{obs}$). Then, we input the sample, along with this stance judgment and rationale, into our \textit{calibration network} (using RoBERTa-base in our setup) to obtain the debiased stance output. We train the \textit{calibration network} using the cross-entropy loss function with ground truth label:
\begin{equation}
    \mathcal{L}_{CE} = -\sum_{i=1}^{N} y_i \log(f_{Cal}(P_{obs}(y_i^{\prime}|\mathcal{I}; x_i, t_i))
\end{equation}

\subsection{Counterfactual Data Augmentation}
\label{subsec:cda}

One challenge in supervised training is the limited representation of bias within the overall training set, and the learned bias features are difficult to generalize. 
To fully leverage our analysis of the existing stance biases in LLMs (Section~\ref{bias-llms}) and facilitate the \textit{calibration network} in learning diverse bias patterns, we generate non-causal and causal \textbf{C}ounterfactual \textbf{A}ugmented \textbf{D}ata (CAD) based on the training data. 

The objective of non-causal counterfactual data augmentation is to explicitly perturb bias-inducing features in the data, enabling the model to identify and mitigate biases. As illustrated in Figure~\ref{Fig:framework} (a), we construct counterfactual samples by modifying non-causal features while preserving stance labels. 
Specifically, we construct an instruction that allows the LLMs to perturb the sentence and target.
To address \textbf{sentiment-stance spurious correlations}, we perturb the text $x_i$ by altering sentiment-related expressions while maintaining the stance. Similarly, to counteract \textbf{target preference bias}, we modify the target $t_i$ while ensuring the stance remains unchanged.
Refer to Figure~\ref{fig-prompt-non-causal} for the detailed prompts.
This obtains the perturbed text $x_i^*$ and perturbed target $t_i^*$. Since we only disturbed confounding, the stance label remains unaffected.
We construct cross-entropy loss on non-causal counterfactual augmented data as follows:
\begin{equation}
\scalebox{0.95}{
$\displaystyle
    \mathcal{L}_{CAD}^\textit{n-cau} = -\sum_{i=1}^{N} y_i \log(f_{Cal}(P_{obs}(y_i^{\prime}|\mathcal{I}; x_i^*, t_i^*))
$}
\end{equation}
In contrast, causal counterfactual data augmentation directly manipulates causal features by reversing the stance. This augmentation improves the model’s capacity to focus on causal features, thereby reducing its reliance on spurious correlations that may introduce bias. As illustrated in Figure~\ref{Fig:framework} (b), we make necessary alterations to text $x_i$ to reverse the stance to target $t_i$, thereby only perturbing the causal features. Refer to Figure~\ref{fig-prompt-causal} for the detailed prompts.
This obtains the perturbed text $\widetilde{x_i}$ expressing a reversed stance to target $t_i$.
We construct cross-entropy loss on causal counterfactual augmented data as follows:
\begin{equation}
\scalebox{0.95}{
$\displaystyle
    \mathcal{L}_{CAD}^\textit{cau} = \sum_{i=1}^{N} y_i \log(f_{Cal}(P_{obs}(y_i^{\prime}|\mathcal{I}; \widetilde{x_i}, t_i)))
$}
\end{equation}

\subsection{Training Objective}

The final training objective incorporates counterfactual augmented data and performs joint training:
\begin{equation}
    \mathcal{L} = \mathcal{L}_{CE} + \mathcal{L}_{CAD}^\textit{n-cau} + \mathcal{L}_{CAD}^\textit{cau}
\end{equation}

\section{Experimental Setup}

\subsection{Datasets}
We conduct experiments of in-target and zero-shot stance detection on three benchmark datasets: SemEval-2016 Task6 (Sem16)~\cite{DBLP:conf/semeval/MohammadKSZC16}, P-Stance~\cite{DBLP:conf/acl/LiSSNIC21} and Varied Stance Topics (VAST)~\cite{DBLP:conf/emnlp/AllawayM20}. The statistic of datasets is shown in Table~\ref{Tab:dataset-statistics}.

\begin{table}[!t]
    \small
    \centering
    \begin{tabular}{ccccc}
        \hline
        Dataset                   & Target                       & Favor & Against & Neutral  \\
        \hline
        \multirow{6}{*}{Sem16}    & HC      & 163   & 565     & 256     \\
                                  & FM      & 268   & 511     & 170     \\
                                  & LA      & 167   & 544     & 222     \\
                                  & A       & 124   & 464     & 145     \\
                                  & CC      & 335   & 26      & 203     \\
                                  & DT      & 148   & 299     & 260     \\
        \hline
        \multirow{3}{*}{P-Stance} & Biden   & 3217  & 4079    & -     \\
                                  & Sanders & 3551  & 2774    & -     \\
                                  & Trump   & 3663  & 4290    & -     \\
        \hline
        VAST                      & -       & 6952  & 7297    & 4296  \\
        \hline
    \end{tabular}
    \caption{Statistics of SemEval-2016 Task6, P-Stance and VAST datasets.}
    \label{Tab:dataset-statistics}
\end{table}

\begin{table*}[t]
\small
\centering
\setlength{\tabcolsep}{3.5pt}
\renewcommand{\arraystretch}{1}
\begin{tabular}{lccccccccccc}
\hline
                                          & \multicolumn{6}{c}{Sem16(\%)}                                                                                                               &                      & \multicolumn{4}{c}{P-Stance(\%)}                                                              \\ \cline{2-7}  \cline{9-12} 
                                          & HC                   & FM                   & LA                   & A                    & CC                   & Avg                  &                      & Biden                & Sanders              & Trump                & Avg                  \\
                                          \hline
\multicolumn{12}{l}{\textbf{Fine-tuning Based Methods}} \\
Roberta                & 55.97          & 68.19          & 67.60          & 65.40          & 43.08          & 58.71          &  & 84.29          & 79.56          & 82.70          & 82.18          \\
BERTweet               & 62.31          & 64.20          & 64.14          & 68.12          & 41.30          & 57.99          &  & 78.09          & 81.02          & 82.48          & 80.53          \\
KPT                    & 71.30          & 63.30          & 63.50          & -              & -              & -              &  & 80.40          & 77.10          & 80.20          & 79.23          \\
KEprompt               & 77.10$^\sharp$          & 68.30$^\sharp$          & 70.30$^\sharp$          & -              & -              & -              &  & 84.40$^\sharp$          & -              & 83.20$^\sharp$          & -              \\
WS-BERT-Dual           & 75.26$^\dag$          & 66.02$^\dag$          & 70.42$^\dag$          & 71.57$^\dag$          & 57.31$^\dag$          & 68.12$^\dag$          &  & 83.50$^\flat$          & 79.00$^\flat$          & 85.80$^\flat$          & 82.77$^\flat$          \\
KASD-BERT              & 77.60$^\dag$          & 70.38$^\dag$          & 72.29$^\dag$          & 72.32$^\dag$          & 61.47$^\dag$          & 70.81$^\dag$          &  & 85.66$^\dag$          & 80.39$^\dag$          & 85.35$^\dag$          & 83.80$^\dag$          \\
\hline
\multicolumn{12}{l}{\textbf{LLaMA-2 Based Methods}}                                                                                                                                          \\
LLaMA-2-Task-Des       & 75.96          & 66.60          & 61.68          & 53.40          & 73.56          & 66.24          &  & 84.31          & 77.29          & 78.08          & 79.89          \\
LLaMA-2-CoT-Demo       & 74.84          & 71.45          & 62.67          & 57.58          & 73.26          & 67.96          &  & 85.03          & 79.77          & 77.52          & 80.77          \\
KASD-LLaMA-2           & 77.89$^\dag$          & 67.29$^\dag$          & 52.00$^\dag$          & 35.78$^\dag$             & 47.12$^\dag$              & 56.02$^\dag$              &  & 79.59$^\dag$          & 71.32$^\dag$          & 67.89$^\dag$          & 72.93$^\dag$          \\
LLaMA-2-7b-FT    & \textbf{81.86} & 71.58          & 65.56 & 68.74          & 75.59          & 72.67          &  & 85.79          & 81.25          & \textbf{87.47} & 84.84          \\
\cdashline{1-12}[2pt/3pt]
\rowcolor[gray]{0.92} FACTUAL$_{\text{LLaMA-2}}$ (Ours)         & 80.44          & \textbf{73.46}$^\star$ & \textbf{67.18}          & \textbf{71.85} & \textbf{76.19} & \textbf{73.82}$^\star$ &  & \textbf{86.34} & \textbf{83.06}$^\star$ & 85.58          & \textbf{84.99} \\
\rowcolor[gray]{0.92} - w/o CAD              & 78.00          & 70.82          & 65.57          & 71.40          & 72.24          & 71.61          &  & 85.35          & 82.00          & 85.51          & 84.29          \\
\hline
\multicolumn{12}{l}{\textbf{GPT-3.5-Turbo Based Methods}}                                                                                                                                    \\
GPT-3.5-Turbo-Task-Des & 73.33          & 66.81          & 67.22          & 25.18          & 72.54          & 61.02          &  & 83.20          & 80.02          & 81.66          & 81.62          \\
GPT-3.5-Turbo-CoT-Demo & 81.58          & 73.42          & 68.28          & 64.96          & 78.35          & 73.32          &  & 83.07          & 77.98          & 81.59          & 80.88          \\
KASD-ChatGPT           & 80.92$^\dag$          & 70.37$^\dag$          & 63.26$^\dag$          & 61.92$^\dag$          & 62.72$^\dag$          & 67.84$^\dag$          &  & 84.59$^\dag$          & 79.96$^\dag$          & \textbf{85.06}$^\dag$          & 83.20$^\dag$          \\
\cdashline{1-12}[2pt/3pt]
\rowcolor[gray]{0.92} FACTUAL$_{\text{GPT-3.5}}$ (Ours)     & \textbf{83.38}$^\star$ & \textbf{78.46}$^\star$ & \textbf{69.36}$^\star$ & \textbf{69.56}$^\star$ & \textbf{80.05}$^\star$ & \textbf{76.16}$^\star$ &  & \textbf{86.03}$^\star$ & \textbf{81.60}$^\star$ & 84.95          & \textbf{84.20}$^\star$ \\
\rowcolor[gray]{0.92} - w/o CAD              & 82.38          & 73.80          & 63.65          & 69.21          & 62.93          & 70.39          &  & 85.40          & 81.36          & 85.00    & 83.92          \\
\hline
\end{tabular}
\caption{In-target stance detection experiment results on Sem16 and P-Stance datasets. The results with $\sharp$ are retrieved from~\cite{huang2023knowledge}, $\flat$ from~\cite{DBLP:conf/wassa/HeML22}, $\dag$ from~\cite{DBLP:conf/emnlp/LiLZZYX23}. The best scores over the same type are in bold. Results with $\star$ indicate significance of our \texttt{FACTUAL} over the same type baseline models at p < 0.05.}
\label{Tab:in-target-results}
\end{table*}

\subsection{Implementation Details}
For GPT-3.5-turbo, we utilize GPT-3.5-turbo-0125. For LLaMA2-70b, we utilize LLaMA2-70b-chat.
We set the temperature to 1.0, top p to 1.0, max tokens to 1024, and fixed the decoding seed to ensure the reproducibility of our experiments.
For our calibration network, we employ the RoBERTa-base model~\cite{DBLP:journals/corr/abs-1907-11692}.
For our counterfactual data augmentation, we employ GPT-3.5-turbo-0301 to generate counterfactual samples, guided by the instructions detailed in Appendix~\ref{sec:prompt_setting}.
We use AdamW as an optimizer with a batch size of 32. Learning rate is set to 1e-5 and weight decay is set to 1e-3. All training was conducted on NVIDIA A100 40G GPUs. We report averaged scores of 5 runs to obtain statistically stable results.

\subsection{Evaluation Metric}
Across three datasets, we used the same evaluation metric established by their proposers, which was also adopted by most of the subsequent baselines.
We adopt the macro-average of the F1-score as the evaluation metric.
For Sem16 and P-Stance, we report $F1 = (F_{favor} + F_{against}) / 2$. For VAST, we report $F1 = (F_{favor} + F_{against} + F_{none}) / 3$.
In the in-target stance detection setting, the model is trained and tested on the same set of targets. We follow the dataset splits provided by the original dataset publisher to ensure a fair comparison with related baselines. The zero-shot stance detection setting presents the model with unseen targets during testing, requiring it to generalize from known targets to infer stances toward new ones. For Sem16 and P-Stance, we take one target as the test set while splitting the remaining targets into training and validation sets in a 7:1 ratio. For VAST, which is inherently a zero-shot stance detection dataset, we use the original splits provided by the dataset publisher. These settings are consistent with the baselines to ensure fairness in comparison.

\subsection{Comparison Models}
The fine-tuned model baselines include vanilla RoBERTa~\cite{DBLP:journals/corr/abs-1907-11692}, domain pre-trained model: BERTweet~\cite{DBLP:conf/emnlp/NguyenVN20}, prompt tuning method KPT~\cite{DBLP:conf/emnlp/ShinRLWS20}, joint contrastive learning framework: JointCL~\cite{DBLP:conf/acl/LiangZL000X22}, incorporating ConceptGraph knowledge model: KEprompt~\cite{huang2023knowledge}, incorporating Wikipedia knowledge model: TarBK-BERT~\cite{DBLP:conf/sigir/ZhuLSDZX22} and WS-BERT~\cite{DBLP:conf/wassa/HeML22}, incorporating knowledge from LLMs: KASD-BERT~\cite{DBLP:conf/emnlp/LiLZZYX23}. 
For large language models, we compare baselines include Task-Des~\cite{DBLP:journals/corr/abs-2212-14548}, CoT-Demo~\cite{DBLP:journals/corr/abs-2304-03087}, the self-consistent chain-of-thought: CoT-SC~\cite{DBLP:conf/iclr/0002WSLCNCZ23}, incorporating Wikipedia knowledge for retrieval-augmented generation: KASD-ChatGPT and KASD-LLaMA-2~\cite{DBLP:conf/emnlp/LiLZZYX23}, fine tuning LLaMA-2-7b using LoRA with training set: LLaMA-2-7b-FT, utilizing collaborative role-infused LLM-based agents: COLA~\cite{DBLP:journals/corr/abs-2310-10467} and utilizing logically consistent chain-of-thought: LC-CoT~\cite{DBLP:journals/corr/abs-2312-16054}\footnote{Since JointCL, TarBK-BERT, COLA, and LC-CoT were only proposed in zero-shot stance detection scenarios, our comparisons with them are conducted solely within the corresponding experimental settings.}.

\begin{table*}[!t]
\small
\centering
\setlength{\tabcolsep}{1.4pt}
\renewcommand{\arraystretch}{1}
\begin{tabular}{lcccccccccccccc}
\hline
                       & \multicolumn{7}{c}{Sem16(\%)}                                                                                        &  & \multicolumn{4}{c}{P-Stance(\%)}                                  &  & VAST(\%)       \\ \cline{2-8}  \cline{10-13} \cline{15-15} 
                       & DT             & HC             & FM             & LA             & A              & CC             & Avg            &  & Biden          & Sanders        & Trump          & Avg            &  & All            \\
                       \hline
\multicolumn{15}{l}{\textbf{Fine-tuning Based Methods}}                                                                                                                                                                                  \\
Roberta                & 32.12          & 43.45          & 40.38          & 38.79          & 26.80          & 18.70          & 33.37          &  & 76.29          & 72.07          & 67.56          & 71.97          &  & 73.18          \\
BERTweet               & 26.88          & 44.82          & 21.97          & 31.91          & 30.49          & 12.48          & 28.09          &  & 73.13          & 68.22          & 67.66          & 69.67          &  & 71.10          \\
JointCL                & 50.50$^\natural$          & 54.80$^\natural$          & 53.80$^\natural$          & 49.50$^\natural$          & 54.50$^\natural$          & 39.70$^\natural$          & 50.47$^\natural$          &  & -              & -              & -              & -              &  & 72.30          \\
TarBK-BERT             & 50.80$^\sharp$          & 55.10$^\sharp$          & 53.80$^\sharp$          & 48.70$^\sharp$          & 56.20$^\sharp$          & 39.50$^\sharp$          & 50.68$^\sharp$          &  & 75.49          & 70.45          & 65.80          & 70.58$^\sharp$          &  & 73.60$^\sharp$          \\
KASD-BERT              & 54.74$^\dag$          & 64.78$^\dag$          & 57.13$^\dag$          & 51.63$^\dag$          & 55.97$^\dag$          & 40.11$^\dag$          & 54.06$^\dag$          &  & 79.04$^\dag$          & 75.09$^\dag$          & 70.84$^\dag$          & 74.99$^\dag$          &  & 76.82$^\dag$          \\
\hline
\multicolumn{15}{l}{\textbf{LLaMA-2 Based Methods}}                                                                                                                                                                                      \\
LLaMA-2-Task-Des       & 66.03          & 73.79          & 71.03          & 66.00          & \textbf{60.44} & 61.91          & 66.53          &  & 82.81          & 78.00          & \textbf{78.87} & 79.89          &  & 68.54          \\
LLaMA-2-CoT-Demo       & 58.56          & 72.09          & 73.83          & 66.10          & 57.58          & 62.47          & 65.11          &  & 83.97          & 79.26          & 77.96          & 80.40          &  & 67.28          \\
KASD-LLaMA-2           & -              & \textbf{77.70}$^\dag$ & 65.57$^\dag$          & 57.07$^\dag$          & 39.55$^\dag$              & 50.72$^\dag$              & -              &  & 75.28$^\dag$          & 74.09$^\dag$          & 69.27$^\dag$          & 72.88$^\dag$          &  & 43.42$^\dag$          \\
LLaMA-2-7b-FT    & 63.99          & 55.49          & 59.46          & 33.18          & 46.37          & 58.24          & 52.79          &  & 83.93          & 77.00          & 74.35          & 78.43          &  & 77.80          \\
\cdashline{1-12}[2pt/3pt]
\rowcolor[gray]{0.92} FACTUAL$_{\text{LLaMA-2}}$ (Ours)         & \textbf{66.96} & 77.19          & \textbf{74.71} & \textbf{72.49}$^\star$ & 58.29          & \textbf{67.71}$^\star$ & \textbf{69.56}$^\star$ &  & \textbf{84.04} & \textbf{81.22}$^\star$ & 77.57          & \textbf{80.94} &  & \textbf{79.62}$^\star$ \\
\rowcolor[gray]{0.92} - w/o CAD              & 61.99          & 69.22          & 62.77          & 60.39          & 40.83          & 63.69          & 59.81          &  & 83.09          & 78.21          & 76.74          & 79.35          &  & 76.61          \\
\hline
\multicolumn{15}{l}{\textbf{GPT-3.5-Turbo Based Methods}}                                                                                                                                                                                \\
GPT-3.5-Turbo-Task-Des & 61.72          & 72.70          & 71.71          & 67.89          & 28.87          & 59.36          & 60.38          &  & 84.08          & 80.38          & 82.38          & 82.28          &  & 50.21          \\
GPT-3.5-Turbo-CoT-Demo & 64.16          & 78.69          & 73.22          & 72.84          & 65.15          & \textbf{75.20} & 71.54          &  & 84.08          & 80.12          & 82.24          & 82.15          &  & 70.14          \\
KASD-ChatGPT           & 64.23$^\dag$          & 80.32$^\dag$          & 70.41$^\dag$          & 62.71$^\dag$          & 63.95$^\dag$          & 55.83$^\dag$          & 66.24$^\dag$          &  & 83.60$^\dag$          & 79.66$^\dag$          & 84.31$^\dag$          & 82.52$^\dag$          &  & 67.03$^\dag$          \\
COLA                   & 71.20$^\ddagger$          & 75.90$^\ddagger$          & 69.10$^\ddagger$          & \textbf{71.00}$^\ddagger$          & 62.30$^\ddagger$          & 64.00$^\ddagger$          & 68.92$^\ddagger$          &  & -              & -              & -              & -              &  & 73.40$^\ddagger$          \\
LC-CoT                 & 71.70$^\flat$          & \textbf{82.90}$^\flat$ & 70.40$^\flat$          & 63.20$^\flat$          & -              & -              & -              &  & -              & -              & -              & -              &  & 72.50$^\flat$          \\
\cdashline{1-12}[2pt/3pt]
\rowcolor[gray]{0.92} FACTUAL$_{\text{GPT-3.5}}$ (Ours)     & \textbf{72.80}$^\star$ & 80.26          & \textbf{75.76}$^\star$ & 68.77$^\star$ & \textbf{66.54}$^\star$ & 71.00          & \textbf{72.52}$^\star$ &  & \textbf{85.14} & \textbf{81.05}$^\star$ & \textbf{85.08} & \textbf{83.76}$^\star$ &  & \textbf{79.98}$^\star$ \\
\rowcolor[gray]{0.92} - w/o CAD              & 63.28          & 72.65          & 60.88          & 62.07          & 41.65          & 67.80          & 61.39          &  & 84.26          & 77.80          & 75.26          & 79.11          &  & 77.50          \\
\hline
\end{tabular}
\caption{Zero-shot stance detection experiment results on Sem16, P-Stance and VAST dataset. The results with $\natural$ are retrieved from~\cite{DBLP:conf/acl/LiangZL000X22}, $\sharp$ from~\cite{DBLP:conf/sigir/ZhuLSDZX22}, $\dag$ from~\cite{DBLP:conf/emnlp/LiLZZYX23}, $\ddagger$ from~\cite{DBLP:journals/corr/abs-2310-10467}, $\flat$ from~\cite{DBLP:journals/corr/abs-2312-16054}. The best scores over the same type are in bold. Results with $\star$ denote the significance tests of our \texttt{FACTUAL} over the same type baseline models at p-value < 0.05.}
\label{Tab:zero-shot-results}
\end{table*}

\section{Experimental Results}
\subsection{In-Target Stance Detection}

We perform experiments on Sem16 and P-Stance for in-target stance detection. The results are presented in Table~\ref{Tab:in-target-results}. It shows that our \texttt{FACTUAL} outperforms all baselines based on different large language models. 
We can observe that \textit{FACTUAL - w/o CAD}, which is without the counterfactual data enhancement, the calibration network trained exclusively on the training set data can still improve stance detection performance.
Moreover, the application of counterfactual data enhancement furthers the model's performance.
When compared to the \textit{LLaMA-2-7b-FT} method, which fine-tunes LLaMA-2-7b, our method attains superior accuracy in stance detection with significantly reduced computation resources.

\subsection{Zero-Shot Stance Detection}

We conduct experiments on Sem16, P-Stance, and VAST for zero-shot stance detection. The results are shown in Table~\ref{Tab:zero-shot-results}. 
It shows that our \texttt{FACTUAL} outperforms all baselines including both fine-tuned models and large language models. 
This indicates that our \texttt{FACTUAL} has strong generalization capabilities and can perform well on unseen targets. 
We can observe that \textit{FACTUAL - w/o CAD} exhibits subpar performance. This can be attributed to the constraints posed by the exclusive reliance on fine-tuning within the limited training dataset, making it an uphill task to generalize the model's debiasing capability. Conversely, employing our counterfactual data enhancement bolsters the out-of-domain generalization prowess of the model considerably, yielding impressive results in zero-shot performance. Compared to the \textit{LLaMA-2-7b-FT}, which performs poorly on zero-shot tasks, \texttt{FACTUAL} demonstrates strong generalization capabilities.

\subsection{Mitigating Biases Effect Analysis}

\begin{table}[t]
\small
\centering
\setlength{\tabcolsep}{1pt}
\renewcommand{\arraystretch}{1}
\begin{tabular}{lcccccccc}
\hline
                       & \multicolumn{2}{c}{Sem16} & \multicolumn{1}{l}{} & \multicolumn{2}{c}{P-Stance} & \multicolumn{1}{l}{} & \multicolumn{2}{c}{VAST} \\
                       \cline{2-3} \cline{5-6} \cline{8-9}
                       & SSC$\downarrow$           & TPB$\downarrow$         &                      & SSC$\downarrow$            & TPB$\downarrow$           &                      & SSC$\downarrow$          & TPB$\downarrow$         \\
\hline
\multicolumn{9}{l}{\textbf{LLaMA-2 Based Methods}}                                                                                                                                             \\
Task-Des       & 17.80              & 17.59            &                      & 23.36               & 9.09               &                      & 23.87             & 17.76            \\
CoT-Demo       & 27.52              & 27.56            &                      & 22.81               & 11.57              &                      & 22.55             & 9.64             \\
CoT-SC             & 33.67              & 27.18            &                      & 29.85               & 6.34               &                      & 31.98             & 23.70            \\
LLaMA-2-7b-FT    & 22.44              & 25.09            &                      & 18.14               & 5.07               &                      & 22.36             & 6.84             \\
KASD-LLaMA-2           & 18.74              & 10.90            &                      & 18.74               & 4.43               &                      & 20.51             & 18.00            \\
\cdashline{1-9}[2pt/3pt]
\rowcolor[gray]{0.92} FACTUAL$_{\text{LLaMA-2}}$         & \textbf{9.61}$^\star$      & \textbf{5.52}$^\star$    &                      & \textbf{17.15}      & \textbf{2.81}$^\star$      &                      & \textbf{19.89}    & \textbf{5.42}    \\
\rowcolor[gray]{0.92} - w/o CAD              & 15.43              & 12.07            &                      & 19.15               & 5.81               &                      & 21.89             & 11.42            \\
\hline
\multicolumn{9}{l}{\textbf{GPT-3.5-Turbo Based Methods}}                                                                                                                                       \\
Task-Des & 27.13              & 22.64            &                      & 23.72               & 5.43               &                      & 28.70             & 28.44            \\
CoT-Demo & 18.08              & 13.47            &                      & 22.75               & 6.61               &                      & 16.32             & 18.40            \\
CoT-SC       & 21.42              & 16.03            &                      & 26.30               & 8.06               &                      & 22.58             & 21.63            \\
KASD-ChatGPT           & 19.49              & 20.17            &                      & 20.90               & 13.94              &                      & 20.54             & 15.56            \\
\cdashline{1-9}[2pt/3pt]
\rowcolor[gray]{0.92} FACTUAL$_{\text{GPT-3.5}}$         & \textbf{11.31}$^\star$     & \textbf{7.74}$^\star$    &                      & \textbf{16.00}$^\star$      & \textbf{3.38}      &                      & \textbf{13.25}$^\star$    & \textbf{7.03}$^\star$    \\
\rowcolor[gray]{0.92} - w/o CAD              & 12.92              & 11.83            &                      & 18.00               & 4.38               &                      & 17.27             & 12.04            \\
\hline
\end{tabular}
\caption{Results of Bias-SSC and Bias-TPB in in-target stance detection on Sem16, P-Stance, and zero-shot stance detection on VAST. The best scores are in bold. Results with $\star$ denote the significance tests of our \texttt{FACTUAL} over the same type baseline models at p-value < 0.05.}
\label{Tab:debias-results}
\end{table}

We conduct experiments to evaluate the Bias-SSC and Bias-TPB of LLMs and further assess the impact of our bias mitigation efforts. The results are shown in Table~\ref{Tab:debias-results}, which indicate that our \texttt{FACTUAL} can effectively alleviate Bias-SSC and Bias-TPB for both GPT-3.5-turbo and LLaMA2-70b, thus validating its effectiveness in mitigating biases. The inclusion of counterfactual data augmentation can effectively improve its debiasing ability, indicating the importance of our counterfactual data augmentation. Our findings also highlight that the integration of counterfactual data augmentation enhances the debiasing capacity of the model, thereby emphasizing the significance of this augmentation in our methodology.

\subsection{Ablation Study}

\begin{table}[t]
\small
\centering
\setlength{\tabcolsep}{1.3pt}
\renewcommand{\arraystretch}{1.1}
\begin{tabular}{lccccccc}
\hline
                   & \multicolumn{3}{c}{Sem16}          &  & \multicolumn{3}{c}{VAST}           \\
                   \cline{2-4} \cline{6-8}
                   & Avg$\uparrow$         & SSC$\downarrow$      & TPB$\downarrow$      &  & All$\uparrow$         & SSC$\downarrow$      & TPB$\downarrow$      \\
\hline
\rowcolor[gray]{0.92} FACTUAL$_{\text{LLaMA-2}}$     & \textbf{73.82}$^\star$ & \textbf{9.61}  & \textbf{5.52}$^\star$ &  & \textbf{79.62} & \textbf{19.89}$^\star$ & 5.42          \\
w/o Calibration    & 67.96          & 27.52          & 27.56         &  & 67.28          & 22.55          & 9.64          \\
\cdashline{1-8}[2pt/3pt]
w/o CAD            & 71.61          & 15.43          & 12.07         &  & 76.61          & 21.89          & 11.42         \\
- w/o non-causal & 71.29          & 13.12          & 13.25         &  & 79.38          & 24.04          & 7.12          \\
- w/o causal     & 69.39          & 10.66          & 14.82         &  & 77.45          & 21.69          & \textbf{3.97} \\
\hline
\rowcolor[gray]{0.92} FACTUAL$_{\text{GPT-3.5}}$     & \textbf{76.16}$^\star$ & 11.31          & \textbf{7.74}$^\star$ &  & \textbf{79.98} & \textbf{13.25}$^\star$ & 7.03          \\
w/o Calibration    & 73.32          & 18.08          & 13.47         &  & 70.14          & 16.32          & 18.40         \\
\cdashline{1-8}[2pt/3pt]
w/o CAD            & 70.39          & 12.92          & 11.83         &  & 77.50          & 17.27          & 12.04         \\
- w/o non-causal & 74.08          & 14.36          & 11.80         &  & 79.38          & 26.48          & 9.32          \\
- w/o causal     & 71.71          & \textbf{10.22} & 10.21         &  & 77.80          & 22.25          & \textbf{6.81} \\
\hline
\end{tabular}
\caption{Experimental results of ablation study of in-target stance detection on the Sem16, and zero-shot stance detection on VAST. Results on P-Stance are shown in Table~\ref{Tab:ablation-results-p-stance}. The best scores are in bold. Results with $\star$ indicate the significance tests of our \texttt{FACTUAL} over the ablation experiments at p-value < 0.05.}
\label{Tab:ablation-results}
\end{table}

We conduct ablation studies to examine the impact of different components in our \texttt{FACTUAL}:
(1) "\textbf{w/o} Calibration" denotes without the calibration network, letting the LLMs directly output stance labels.
(2) "\textbf{w/o} CAD" denotes without the counterfactual augmented data when training the calibration network. 
(3) "\textbf{w/o} non-causal" denotes without the non-causal counterfactual augmented data when training the calibration network. 
(4) "\textbf{w/o} causal" denotes without the causal counterfactual augmented data when training the calibration network. 
 
The results are presented in Table~\ref{Tab:ablation-results}. 
Note that despite utilizing the same stance reasoning, a lack of calibration can result in sub-optimal results and notable biases. Thus validating the effectiveness of our gate calibration network. Analysis in Appendix~\ref{sec:prompt_robustness} demonstrate that our method exhibits robustness across different prompt templates.

Additionally, removing non-causal counterfactual data significantly increases bias, highlighting its crucial role in bias mitigation. Conversely, eliminating causal counterfactual data markedly reduces performance, underscoring its substantial impact on the accuracy and generalizability of the calibration network.

\section{Conclusion}

In this paper, we categorize the biases of LLMs in stance detection into two types from the perspective of causality and propose metrics to quantify these biases.
Then, we propose a Counter\textbf{fact}ual A\textbf{u}gmented C\textbf{al}ibration Network, coined as \texttt{FACTUAL}. In which, a trainable calibration network and counterfactual data augmentation are explored to mitigate the biases of LLMs in stance detection. Experimental results on in-target and zero-shot stance detection show that our \texttt{FACTUAL} can effectively reduce the bias of LLMs in stance detection and contribute to improved performance.

\section*{Acknowledgements}
This work was partially supported by the National Natural Science Foundation of China 62176076,  Natural Science Foundation of Guang Dong 2023A1515012922, the Shenzhen Foundational Research Funding JCYJ20220818102415032, the Major Key Project of PCL2021A06, CIPSC-SMP-ZHIPU Large Model Cross-Disciplinary Fund ZPCG20241119405 and Hong Kong RGC GRF No. 14206324.

\section*{Limitations}
Our framework involves using GPT-3.5 to generate counterfactual augmented data. As we discussed in Appendix~\ref{sec:human_eval}, these samples may contain errors, but overall are beneficial to the training of our calibration network. The methods of constructing counterfactual augmented data using manual annotation or other methods remain to be explored. While our study primarily focuses on stance biases—particularly sentiment-stance spurious correlations and target preference bias—our approach can be extended to mitigate other forms of bias by designing appropriate counterfactual samples.

\section*{Ethics Statement}
The datasets used in this paper are sourced from open-access datasets. The VAST dataset provides complete text data in open access. In compliance with the privacy agreement of Twitter for academic usage, the Sem16 and P-Stance were accessed using the official Twitter API\footnote{https://developer.twitter.com/en/docs/twitter-api} through the Tweet IDs to fetch complete text data. We removed the information on user privacy from the data. In these datasets, we analyze the biases and stereotypes in stance detection for some sensitive targets (e.g., belief, politics, etc.). We DO NOT critique any biases and stereotypes. We focus on analyzing their impacts on stance detection and mitigating these impacts. We used the counterfactual augmented data obtained from the GPT-3.5-Turbo API service from OpenAI. We followed their term and policies. Some examples in our paper may include a stance or tendency. It should be clarified that they are randomly sampled from the dataset for better studying the dataset and task, and do not represent any personal viewpoints.

\bibliography{custom}

\begin{thebibliography}{35}
\providecommand{\natexlab}[1]{#1}

\bibitem[{Allaway and McKeown(2020)}]{DBLP:conf/emnlp/AllawayM20}
Emily Allaway and Kathleen~R. McKeown. 2020.
\newblock \href {https://doi.org/10.18653/V1/2020.EMNLP-MAIN.717} {Zero-shot stance detection: {A} dataset and model using generalized topic representations}.
\newblock In \emph{Proceedings of the 2020 Conference on Empirical Methods in Natural Language Processing, {EMNLP} 2020, Online, November 16-20, 2020}, pages 8913--8931. Association for Computational Linguistics.

\bibitem[{Chen et~al.(2021)Chen, Ye, and Cui}]{DBLP:conf/icann/ChenYC21}
Pengyuan Chen, Kai Ye, and Xiaohui Cui. 2021.
\newblock \href {https://doi.org/10.1007/978-3-030-86365-4\_22} {Integrating n-gram features into pre-trained model: {A} novel ensemble model for multi-target stance detection}.
\newblock In \emph{Artificial Neural Networks and Machine Learning - {ICANN} 2021 - 30th International Conference on Artificial Neural Networks, Bratislava, Slovakia, September 14-17, 2021, Proceedings, Part {III}}, volume 12893 of \emph{Lecture Notes in Computer Science}, pages 269--279. Springer.

\bibitem[{Dong et~al.(2023)Dong, Zhu, Wang, Teleki, and Caverlee}]{DBLP:journals/corr/abs-2310-12490}
Xiangjue Dong, Ziwei Zhu, Zhuoer Wang, Maria Teleki, and James Caverlee. 2023.
\newblock \href {https://doi.org/10.48550/ARXIV.2310.12490} {Co{\textdollar}{\^{}}2{\textdollar}pt: Mitigating bias in pre-trained language models through counterfactual contrastive prompt tuning}.
\newblock \emph{CoRR}, abs/2310.12490.

\bibitem[{Ghosh et~al.(2019)Ghosh, Singhania, Singh, Rudra, and Ghosh}]{DBLP:conf/clef/GhoshSSRG19}
Shalmoli Ghosh, Prajwal Singhania, Siddharth Singh, Koustav Rudra, and Saptarshi Ghosh. 2019.
\newblock \href {https://doi.org/10.1007/978-3-030-28577-7\_4} {Stance detection in web and social media: {A} comparative study}.
\newblock In \emph{Experimental {IR} Meets Multilinguality, Multimodality, and Interaction - 10th International Conference of the {CLEF} Association, {CLEF} 2019, Lugano, Switzerland, September 9-12, 2019, Proceedings}, volume 11696 of \emph{Lecture Notes in Computer Science}, pages 75--87. Springer.

\bibitem[{Gon{\c{c}}alves and Strubell(2023)}]{DBLP:conf/emnlp/GoncalvesS23}
Gustavo Gon{\c{c}}alves and Emma Strubell. 2023.
\newblock \href {https://aclanthology.org/2023.emnlp-main.161} {Understanding the effect of model compression on social bias in large language models}.
\newblock In \emph{Proceedings of the 2023 Conference on Empirical Methods in Natural Language Processing, {EMNLP} 2023, Singapore, December 6-10, 2023}, pages 2663--2675. Association for Computational Linguistics.

\bibitem[{He et~al.(2023)He, Guo, Rao, and Lerman}]{DBLP:journals/corr/abs-2311-09687}
Zihao He, Siyi Guo, Ashwin Rao, and Kristina Lerman. 2023.
\newblock \href {https://doi.org/10.48550/ARXIV.2311.09687} {Inducing political bias allows language models anticipate partisan reactions to controversies}.
\newblock \emph{CoRR}, abs/2311.09687.

\bibitem[{He et~al.(2022)He, Mokhberian, and Lerman}]{DBLP:conf/wassa/HeML22}
Zihao He, Negar Mokhberian, and Kristina Lerman. 2022.
\newblock \href {https://doi.org/10.18653/V1/2022.WASSA-1.7} {Infusing knowledge from wikipedia to enhance stance detection}.
\newblock In \emph{Proceedings of the 12th Workshop on Computational Approaches to Subjectivity, Sentiment {\&} Social Media Analysis, WASSA@ACL 2022, Dublin, Ireland, May 26, 2022}, pages 71--77. Association for Computational Linguistics.

\bibitem[{Huang et~al.(2023)Huang, Zhang, Li, Zhang, Sun, Luo, and Peng}]{huang2023knowledge}
Hu~Huang, Bowen Zhang, Yangyang Li, Baoquan Zhang, Yuxi Sun, Chuyao Luo, and Cheng Peng. 2023.
\newblock Knowledge-enhanced prompt-tuning for stance detection.
\newblock \emph{ACM Transactions on Asian and Low-Resource Language Information Processing}.

\bibitem[{Jang and Allan(2018)}]{DBLP:conf/sigir/JangA18}
Myungha Jang and James Allan. 2018.
\newblock \href {https://doi.org/10.1145/3209978.3210143} {Explaining controversy on social media via stance summarization}.
\newblock In \emph{The 41st International {ACM} {SIGIR} Conference on Research {\&} Development in Information Retrieval, {SIGIR} 2018, Ann Arbor, MI, USA, July 08-12, 2018}, pages 1221--1224. {ACM}.

\bibitem[{Jenny et~al.(2023)Jenny, Billeter, Sachan, Sch{\"{o}}lkopf, and Jin}]{DBLP:journals/corr/abs-2311-08605}
David~F. Jenny, Yann Billeter, Mrinmaya Sachan, Bernhard Sch{\"{o}}lkopf, and Zhijing Jin. 2023.
\newblock \href {https://doi.org/10.48550/ARXIV.2311.08605} {Navigating the ocean of biases: Political bias attribution in language models via causal structures}.
\newblock \emph{CoRR}, abs/2311.08605.

\bibitem[{Kaushal et~al.(2021)Kaushal, Saha, and Ganguly}]{DBLP:conf/naacl/KaushalSG21}
Ayush Kaushal, Avirup Saha, and Niloy Ganguly. 2021.
\newblock \href {https://doi.org/10.18653/V1/2021.NAACL-MAIN.303} {twt-wt: {A} dataset to assert the role of target entities for detecting stance of tweets}.
\newblock In \emph{Proceedings of the 2021 Conference of the North American Chapter of the Association for Computational Linguistics: Human Language Technologies, {NAACL-HLT} 2021, Online, June 6-11, 2021}, pages 3879--3889. Association for Computational Linguistics.

\bibitem[{Lan et~al.(2023)Lan, Gao, Jin, and Li}]{DBLP:journals/corr/abs-2310-10467}
Xiaochong Lan, Chen Gao, Depeng Jin, and Yong Li. 2023.
\newblock \href {https://doi.org/10.48550/ARXIV.2310.10467} {Stance detection with collaborative role-infused llm-based agents}.
\newblock \emph{CoRR}, abs/2310.10467.

\bibitem[{Li et~al.(2023)Li, Liang, Zhao, Zhang, Yang, and Xu}]{DBLP:conf/emnlp/LiLZZYX23}
Ang Li, Bin Liang, Jingqian Zhao, Bowen Zhang, Min Yang, and Ruifeng Xu. 2023.
\newblock \href {https://aclanthology.org/2023.emnlp-main.972} {Stance detection on social media with background knowledge}.
\newblock In \emph{Proceedings of the 2023 Conference on Empirical Methods in Natural Language Processing, {EMNLP} 2023, Singapore, December 6-10, 2023}, pages 15703--15717. Association for Computational Linguistics.

\bibitem[{Li et~al.(2021)Li, Sosea, Sawant, Nair, Inkpen, and Caragea}]{DBLP:conf/acl/LiSSNIC21}
Yingjie Li, Tiberiu Sosea, Aditya Sawant, Ajith~Jayaraman Nair, Diana Inkpen, and Cornelia Caragea. 2021.
\newblock \href {https://doi.org/10.18653/v1/2021.findings-acl.208} {P-stance: {A} large dataset for stance detection in political domain}.
\newblock In \emph{Findings of the Association for Computational Linguistics: {ACL/IJCNLP} 2021, Online Event, August 1-6, 2021}, volume {ACL/IJCNLP} 2021 of \emph{Findings of {ACL}}, pages 2355--2365. Association for Computational Linguistics.

\bibitem[{Liang et~al.(2022)Liang, Zhu, Li, Yang, Gui, He, and Xu}]{DBLP:conf/acl/LiangZL000X22}
Bin Liang, Qinglin Zhu, Xiang Li, Min Yang, Lin Gui, Yulan He, and Ruifeng Xu. 2022.
\newblock \href {https://doi.org/10.18653/V1/2022.ACL-LONG.7} {Jointcl: {A} joint contrastive learning framework for zero-shot stance detection}.
\newblock In \emph{Proceedings of the 60th Annual Meeting of the Association for Computational Linguistics (Volume 1: Long Papers), {ACL} 2022, Dublin, Ireland, May 22-27, 2022}, pages 81--91. Association for Computational Linguistics.

\bibitem[{Limisiewicz et~al.(2023)Limisiewicz, Marecek, and Musil}]{DBLP:journals/corr/abs-2310-18913}
Tomasz Limisiewicz, David Marecek, and Tom{\'{a}}s Musil. 2023.
\newblock \href {https://doi.org/10.48550/ARXIV.2310.18913} {Debiasing algorithm through model adaptation}.
\newblock \emph{CoRR}, abs/2310.18913.

\bibitem[{Liu et~al.(2019)Liu, Ott, Goyal, Du, Joshi, Chen, Levy, Lewis, Zettlemoyer, and Stoyanov}]{DBLP:journals/corr/abs-1907-11692}
Yinhan Liu, Myle Ott, Naman Goyal, Jingfei Du, Mandar Joshi, Danqi Chen, Omer Levy, Mike Lewis, Luke Zettlemoyer, and Veselin Stoyanov. 2019.
\newblock \href {https://arxiv.org/abs/1907.11692} {Roberta: {A} robustly optimized {BERT} pretraining approach}.
\newblock \emph{CoRR}, abs/1907.11692.

\bibitem[{Luo et~al.(2023)Luo, Yang, Meng, Li, Zhou, and Zhang}]{DBLP:journals/corr/abs-2308-08747}
Yun Luo, Zhen Yang, Fandong Meng, Yafu Li, Jie Zhou, and Yue Zhang. 2023.
\newblock \href {https://doi.org/10.48550/ARXIV.2308.08747} {An empirical study of catastrophic forgetting in large language models during continual fine-tuning}.
\newblock \emph{CoRR}, abs/2308.08747.

\bibitem[{Mohammad et~al.(2016)Mohammad, Kiritchenko, Sobhani, Zhu, and Cherry}]{DBLP:conf/semeval/MohammadKSZC16}
Saif~M. Mohammad, Svetlana Kiritchenko, Parinaz Sobhani, Xiaodan Zhu, and Colin Cherry. 2016.
\newblock \href {https://doi.org/10.18653/V1/S16-1003} {Semeval-2016 task 6: Detecting stance in tweets}.
\newblock In \emph{Proceedings of the 10th International Workshop on Semantic Evaluation, SemEval@NAACL-HLT 2016, San Diego, CA, USA, June 16-17, 2016}, pages 31--41. The Association for Computer Linguistics.

\bibitem[{Nguyen et~al.(2020)Nguyen, Vu, and Nguyen}]{DBLP:conf/emnlp/NguyenVN20}
Dat~Quoc Nguyen, Thanh Vu, and Anh~Tuan Nguyen. 2020.
\newblock \href {https://doi.org/10.18653/v1/2020.emnlp-demos.2} {Bertweet: {A} pre-trained language model for english tweets}.
\newblock In \emph{Proceedings of the 2020 Conference on Empirical Methods in Natural Language Processing: System Demonstrations, {EMNLP} 2020 - Demos, Online, November 16-20, 2020}, pages 9--14. Association for Computational Linguistics.

\bibitem[{Salinas et~al.(2023)Salinas, Penafiel, McCormack, and Morstatter}]{DBLP:journals/corr/abs-2310-08780}
Abel Salinas, Louis Penafiel, Robert McCormack, and Fred Morstatter. 2023.
\newblock \href {https://doi.org/10.48550/ARXIV.2310.08780} {"im not racist but...": Discovering bias in the internal knowledge of large language models}.
\newblock \emph{CoRR}, abs/2310.08780.

\bibitem[{Shin et~al.(2020)Shin, Razeghi, IV, Wallace, and Singh}]{DBLP:conf/emnlp/ShinRLWS20}
Taylor Shin, Yasaman Razeghi, Robert L.~Logan IV, Eric Wallace, and Sameer Singh. 2020.
\newblock \href {https://doi.org/10.18653/v1/2020.emnlp-main.346} {Autoprompt: Eliciting knowledge from language models with automatically generated prompts}.
\newblock In \emph{Proceedings of the 2020 Conference on Empirical Methods in Natural Language Processing, {EMNLP} 2020, Online, November 16-20, 2020}, pages 4222--4235. Association for Computational Linguistics.

\bibitem[{Somasundaran and Wiebe(2010)}]{somasundaran-wiebe-2010-recognizing}
Swapna Somasundaran and Janyce Wiebe. 2010.
\newblock \href {https://aclanthology.org/W10-0214} {Recognizing stances in ideological on-line debates}.
\newblock In \emph{Proceedings of the {NAACL} {HLT} 2010 Workshop on Computational Approaches to Analysis and Generation of Emotion in Text}, pages 116--124, Los Angeles, CA. Association for Computational Linguistics.

\bibitem[{Stefanov et~al.(2020)Stefanov, Darwish, Atanasov, and Nakov}]{DBLP:conf/acl/StefanovDAN20}
Peter Stefanov, Kareem Darwish, Atanas Atanasov, and Preslav Nakov. 2020.
\newblock \href {https://doi.org/10.18653/V1/2020.ACL-MAIN.50} {Predicting the topical stance and political leaning of media using tweets}.
\newblock In \emph{Proceedings of the 58th Annual Meeting of the Association for Computational Linguistics, {ACL} 2020, Online, July 5-10, 2020}, pages 527--537. Association for Computational Linguistics.

\bibitem[{Sun et~al.(2018)Sun, Wang, Zhu, and Zhou}]{DBLP:conf/coling/SunWZZ18}
Qingying Sun, Zhongqing Wang, Qiaoming Zhu, and Guodong Zhou. 2018.
\newblock \href {https://aclanthology.org/C18-1203/} {Stance detection with hierarchical attention network}.
\newblock In \emph{Proceedings of the 27th International Conference on Computational Linguistics, {COLING} 2018, Santa Fe, New Mexico, USA, August 20-26, 2018}, pages 2399--2409. Association for Computational Linguistics.

\bibitem[{Touvron et~al.(2023)Touvron, Lavril, Izacard, Martinet, Lachaux, Lacroix, Rozi{\`{e}}re, Goyal, Hambro, Azhar, Rodriguez, Joulin, Grave, and Lample}]{DBLP:journals/corr/abs-2302-13971}
Hugo Touvron, Thibaut Lavril, Gautier Izacard, Xavier Martinet, Marie{-}Anne Lachaux, Timoth{\'{e}}e Lacroix, Baptiste Rozi{\`{e}}re, Naman Goyal, Eric Hambro, Faisal Azhar, Aur{\'{e}}lien Rodriguez, Armand Joulin, Edouard Grave, and Guillaume Lample. 2023.
\newblock \href {https://doi.org/10.48550/arXiv.2302.13971} {Llama: Open and efficient foundation language models}.
\newblock \emph{CoRR}, abs/2302.13971.

\bibitem[{Wang et~al.(2023)Wang, Wei, Schuurmans, Le, Chi, Narang, Chowdhery, and Zhou}]{DBLP:conf/iclr/0002WSLCNCZ23}
Xuezhi Wang, Jason Wei, Dale Schuurmans, Quoc~V. Le, Ed~H. Chi, Sharan Narang, Aakanksha Chowdhery, and Denny Zhou. 2023.
\newblock \href {https://openreview.net/pdf?id=1PL1NIMMrw} {Self-consistency improves chain of thought reasoning in language models}.
\newblock In \emph{The Eleventh International Conference on Learning Representations, {ICLR} 2023, Kigali, Rwanda, May 1-5, 2023}. OpenReview.net.

\bibitem[{Yuan et~al.(2022{\natexlab{a}})Yuan, Zhao, Lu, and Qin}]{DBLP:conf/coling/YuanZL022}
Jianhua Yuan, Yanyan Zhao, Yanyue Lu, and Bing Qin. 2022{\natexlab{a}}.
\newblock \href {https://aclanthology.org/2022.coling-1.596} {{SSR:} utilizing simplified stance reasoning process for robust stance detection}.
\newblock In \emph{Proceedings of the 29th International Conference on Computational Linguistics, {COLING} 2022, Gyeongju, Republic of Korea, October 12-17, 2022}, pages 6846--6858. International Committee on Computational Linguistics.

\bibitem[{Yuan et~al.(2022{\natexlab{b}})Yuan, Zhao, and Qin}]{DBLP:journals/corr/abs-2212-10392}
Jianhua Yuan, Yanyan Zhao, and Bing Qin. 2022{\natexlab{b}}.
\newblock \href {https://doi.org/10.48550/ARXIV.2212.10392} {Debiasing stance detection models with counterfactual reasoning and adversarial bias learning}.
\newblock \emph{CoRR}, abs/2212.10392.

\bibitem[{Zhang et~al.(2022)Zhang, Ding, and Jing}]{DBLP:journals/corr/abs-2212-14548}
Bowen Zhang, Daijun Ding, and Liwen Jing. 2022.
\newblock \href {https://doi.org/10.48550/ARXIV.2212.14548} {How would stance detection techniques evolve after the launch of chatgpt?}
\newblock \emph{CoRR}, abs/2212.14548.

\bibitem[{Zhang et~al.(2023{\natexlab{a}})Zhang, Ding, Jing, and Huang}]{DBLP:journals/corr/abs-2312-16054}
Bowen Zhang, Daijun Ding, Liwen Jing, and Hu~Huang. 2023{\natexlab{a}}.
\newblock \href {https://doi.org/10.48550/ARXIV.2312.16054} {A logically consistent chain-of-thought approach for stance detection}.
\newblock \emph{CoRR}, abs/2312.16054.

\bibitem[{Zhang et~al.(2023{\natexlab{b}})Zhang, Fu, Ding, Huang, Li, and Jing}]{DBLP:journals/corr/abs-2304-03087}
Bowen Zhang, Xianghua Fu, Daijun Ding, Hu~Huang, Yangyang Li, and Liwen Jing. 2023{\natexlab{b}}.
\newblock \href {https://doi.org/10.48550/ARXIV.2304.03087} {Investigating chain-of-thought with chatgpt for stance detection on social media}.
\newblock \emph{CoRR}, abs/2304.03087.

\bibitem[{Zheng et~al.(2023)Zheng, Zhou, Meng, Zhou, and Huang}]{DBLP:journals/corr/abs-2309-03882}
Chujie Zheng, Hao Zhou, Fandong Meng, Jie Zhou, and Minlie Huang. 2023.
\newblock \href {https://doi.org/10.48550/ARXIV.2309.03882} {Large language models are not robust multiple choice selectors}.
\newblock \emph{CoRR}, abs/2309.03882.

\bibitem[{Zhou et~al.(2023)Zhou, Xu, Liu, An, Ai, and Huang}]{DBLP:journals/corr/abs-2311-08648}
Yuhang Zhou, Paiheng Xu, Xiaoyu Liu, Bang An, Wei Ai, and Furong Huang. 2023.
\newblock \href {https://doi.org/10.48550/ARXIV.2311.08648} {Explore spurious correlations at the concept level in language models for text classification}.
\newblock \emph{CoRR}, abs/2311.08648.

\bibitem[{Zhu et~al.(2022)Zhu, Liang, Sun, Du, Zhou, and Xu}]{DBLP:conf/sigir/ZhuLSDZX22}
Qinglin Zhu, Bin Liang, Jingyi Sun, Jiachen Du, Lanjun Zhou, and Ruifeng Xu. 2022.
\newblock \href {https://doi.org/10.1145/3477495.3531807} {Enhancing zero-shot stance detection via targeted background knowledge}.
\newblock In \emph{{SIGIR} '22: The 45th International {ACM} {SIGIR} Conference on Research and Development in Information Retrieval, Madrid, Spain, July 11 - 15, 2022}, pages 2070--2075. {ACM}.

\end{thebibliography}

\appendix

\section{Bias Measurement Analysis}
\label{sec:measurement_analysis}
The standard deviation of recalls (RStd) can largely mitigate the impact of label imbalance in the dataset and effectively assess model bias. This metric is commonly used to evaluate model bias and, as highlighted in relevant research~\cite{DBLP:journals/corr/abs-2309-03882}, "This measurement is intuitive that greater recall imbalance indicates more pronounced selection bias and is not as susceptible to label imbalance as the counting-based measurement."

To illustrate the robustness of this metric against label imbalance, we provide an example from our experiments with the Sem16 dataset. The ground truth data distribution is shown in Table~\ref{Tab:distribution-sem16}. We observed a certain degree of imbalance in label distribution. Based on this, we sampled 200 instances in which ground truth is favor, none, and against, respectively on each sentiment. The sampled instances exhibited no imbalance. We calculated Bias-SSC on these sampled instances, obtaining a result of \textbf{27.34}, compared to \textbf{27.13} on the original dataset. Thus, we believe that the actual distribution of sentiment, target, and stance labels, does not significantly affect our measurement of stance detection bias in LLMs.

\begin{table}[h]
\small
\centering
\setlength{\tabcolsep}{3pt}
\renewcommand{\arraystretch}{1.1}
\begin{tabular}{l|ccc}
\hline
(\%) & Positive & Neutral & Negative \\
\hline
Favor   &       9.96 &      1.99 &      12.79 \\
None    &       7.52 &      2.96 &      15.32 \\
Against &      13.82 &      1.46 &      34.19 \\
\hline
\end{tabular}
\caption{The ground truth data distribution of the Sem16 dataset.}
\label{Tab:distribution-sem16}
\end{table}

\section{Bias Influence}
\label{sec:bias_influence}
Bias represents a macro-level error pattern. In our experiments, the macroscopic results indicate that larger bias-SSC and bias-TPB tend to lead to poorer stance detection results. To prove this, we conducted a comprehensive analysis involving the calculation of Pearson correlation coefficients between bias-SSC and stance detection F1 scores, as well as bias-TPB and stance detection F1 scores, across 97 groups derived from Tables~\ref{Tab:sent-spurious-result}-\ref{Tab:target-prefer-result},~\ref{Tab:debias-results},~\ref{Tab:ablation-results},~\ref{Tab:ablation-results-p-stance},~\ref{Tab:sem16-sentiment-bias}-\ref{Tab:pstance-vast-target-bias} (with bias-SSC and bias-TPB in Tables~\ref{Tab:debias-results},~\ref{Tab:ablation-results},~\ref{Tab:ablation-results-p-stance} and their corresponding F1 values in Tables~\ref{Tab:in-target-results}, \ref{Tab:zero-shot-results}, and bias-SSC and F1 or bias-TPB and F1 values in Tables~\ref{Tab:sem16-sentiment-bias}-\ref{Tab:pstance-vast-target-bias}). The results are shown in Table~\ref{Tab:correlation_coefficients}, which indicate significant negative correlations for bias-SSC and stance detection F1, and for bias-TPB and stance detection F1.

\begin{table}[t]
\small
\centering
\setlength{\tabcolsep}{3pt}
\renewcommand{\arraystretch}{1.1}
\begin{tabular}{lcc}
\hline
Bias Type & Correlation Coefficient & p-value \\
\hline
bias-SSC   &   -0.4047 &   3.9e-05 \\
bias-TPB   &   -0.5323 &   2.0e-08 \\
\hline
\end{tabular}
\caption{Results of Pearson correlation coefficients and p-value between bias-SSC and stance detection F1 scores, as well as bias-TPB and stance detection F1 scores.}
\label{Tab:correlation_coefficients}
\end{table}

\section{Prompts Setting}
\label{sec:prompt_setting}

We present the prompt templates used in Section~\ref{subsec:exp_result} and Section~\ref{subsec:cda}.

\begin{figure*}[t]
\centering  
\includegraphics[width=1\linewidth]{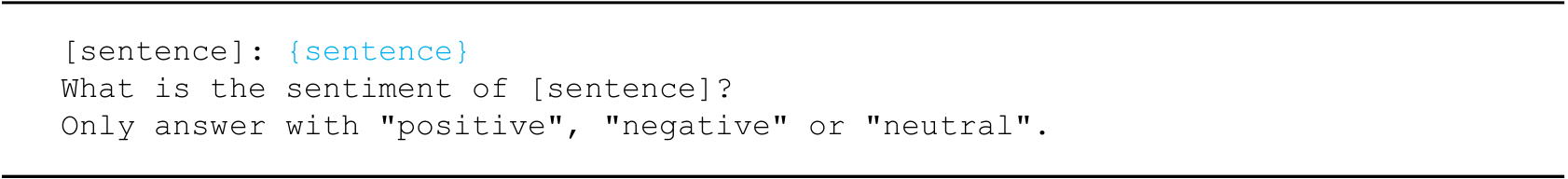}
\caption{Prompt template of sentiment labels annotation by GPT-4. Fill the blue text with the corresponding text from the sample.}
\label{fig-prompt-gpt4-sentiment}
\end{figure*}

\begin{figure*}[t]
\centering  
\includegraphics[width=1\linewidth]{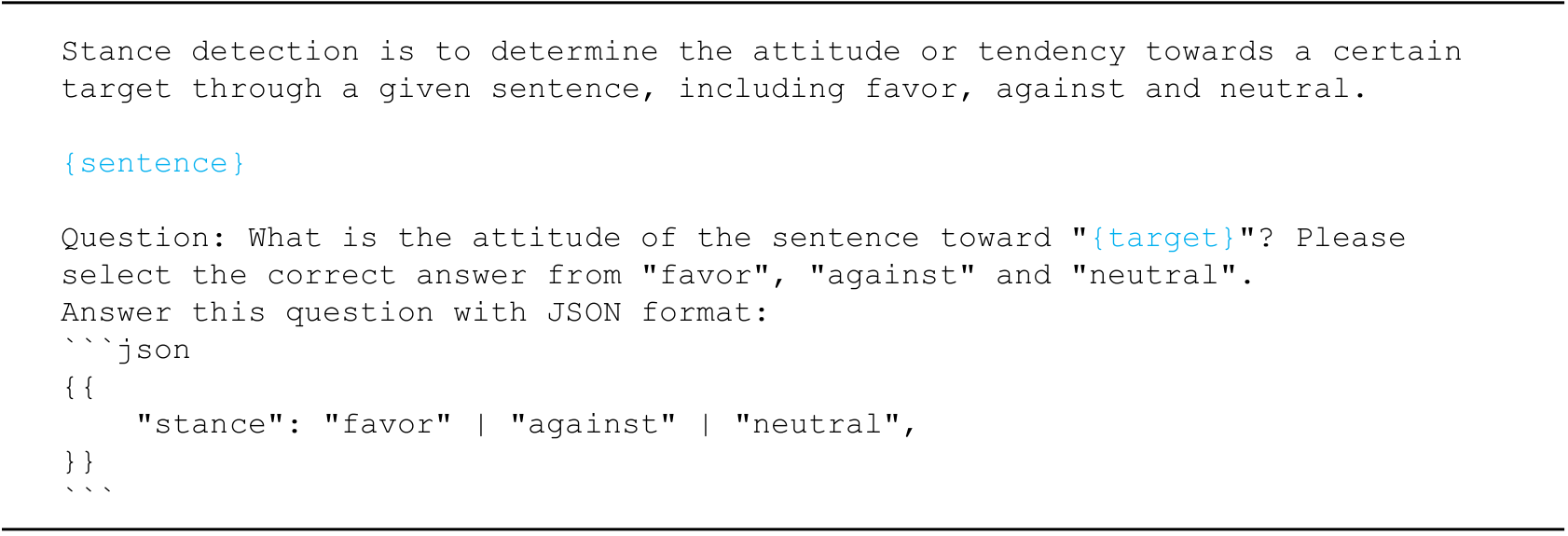}
\caption{Prompt template with Task-Des setting. We first outline the stance detection task, then instruct the LLMs to determine the stance based on the sentence in relation to the target. Fill the blue text with the corresponding text and target from the sample.}
\label{fig-prompt-normal}
\end{figure*}

\begin{figure*}[t]
\centering  
\includegraphics[width=1\linewidth]{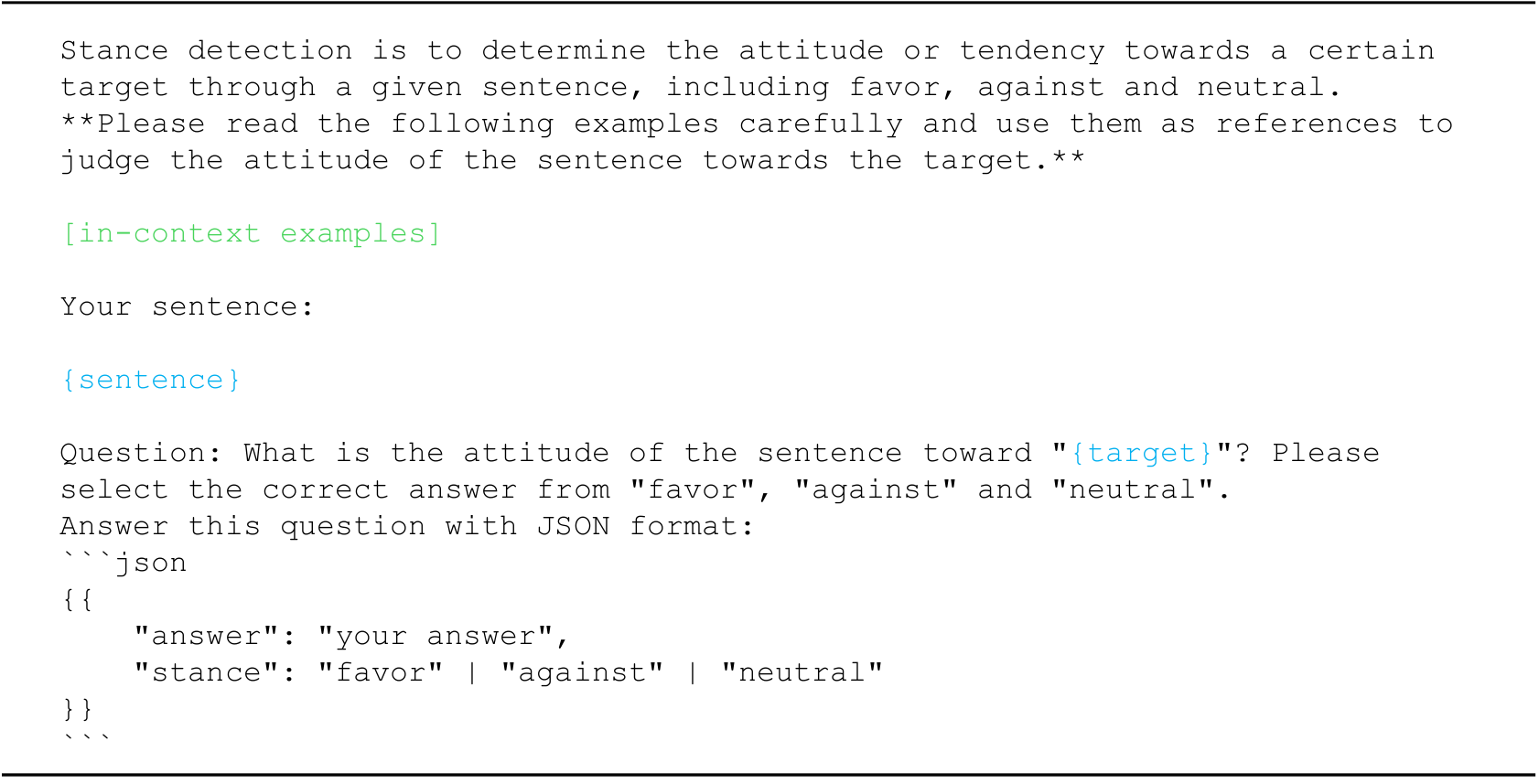}
\caption{Prompt template with CoT-Demo setting. We randomly select 4 samples from the training set, provide the ground truth stance labels, and guide GPT-4 to generate chain-of-thought rationales as examples for this prompt. Fill the green text with constructed examples, and fill the blue text with the corresponding text and target from the sample.}
\label{fig-prompt-4-shot}
\end{figure*}

\begin{figure*}[t]
\centering  
\includegraphics[width=1\linewidth]{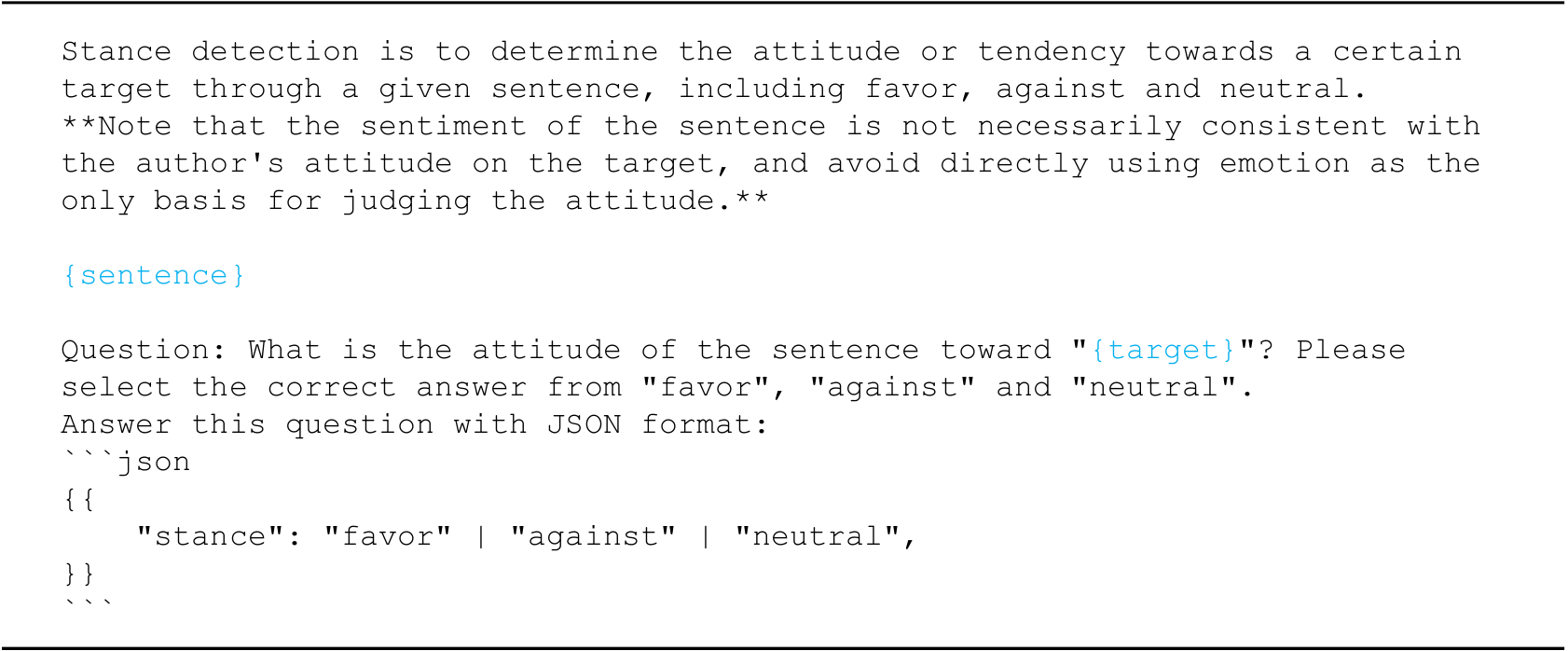}
\caption{Prompt template with SSC Debias-Instruct setting. We add explicit debiasing instructions following the task description. Fill the blue text with the corresponding text and target from the sample.}
\label{fig-prompt-sentiment-debias}
\end{figure*}

\begin{figure*}[t]
\centering  
\includegraphics[width=1\linewidth]{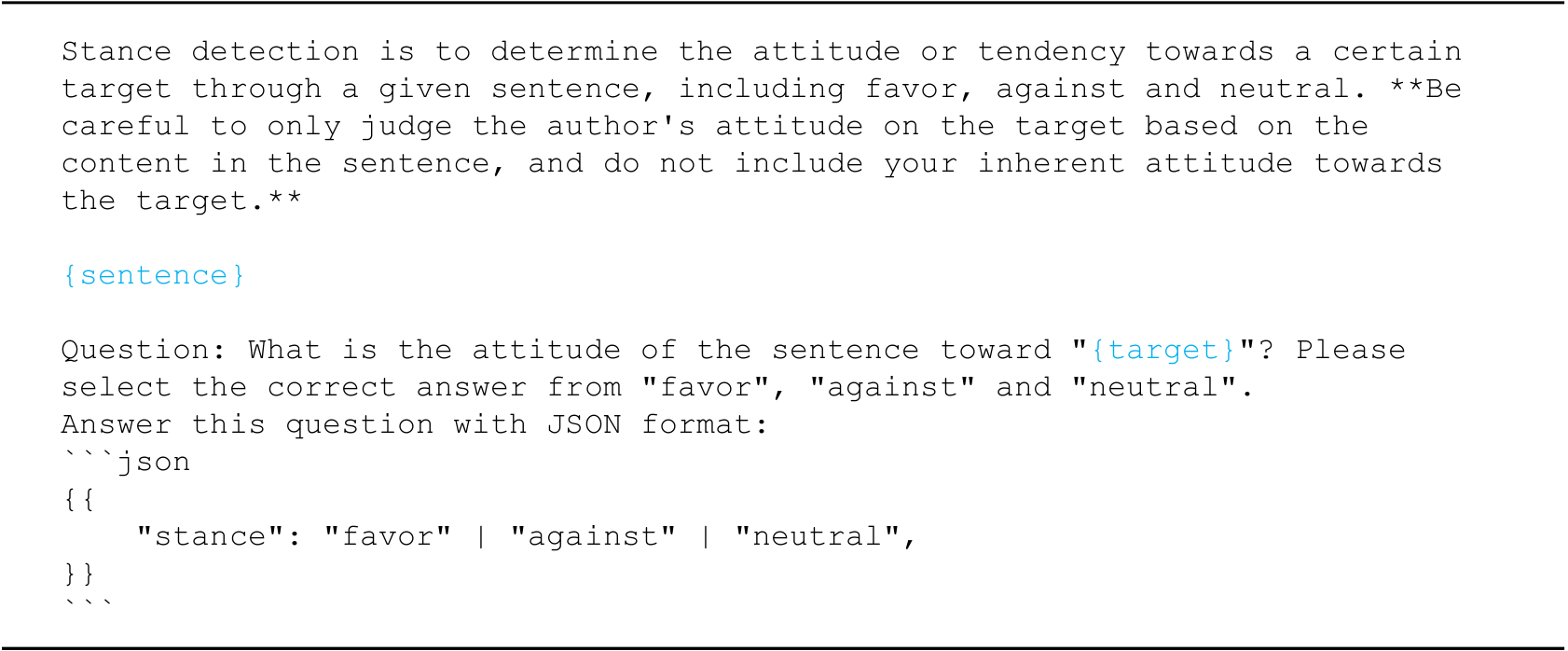}
\caption{Prompt template with TPB Debias-Instruct setting. We add explicit debiasing instructions following the task description. Fill the blue text with the corresponding text and target from the sample.}
\label{fig-prompt-target-debias}
\end{figure*}

Specifically, Figure~\ref{fig-prompt-gpt4-sentiment} shows the prompt template we use with GPT-4 to obtain sentiment labels. Figures~\ref{fig-prompt-normal},~\ref{fig-prompt-4-shot}, and~\ref{fig-prompt-sentiment-debias} show the prompt templates corresponding to the Task-Des, CoT-Demo, and Debias-Instruct prompt settings in Section~\ref{subsubsec: SSC}, respectively. Similarly, Figures~\ref{fig-prompt-normal},~\ref{fig-prompt-4-shot}, and~\ref{fig-prompt-target-debias} display the prompt templates corresponding to the Task-Des, CoT-Demo, and Debias-Instruct prompt settings in Section~\ref{subsubsec: TPB}, respectively.

\begin{figure*}[t]
\centering  
\includegraphics[width=1\linewidth]{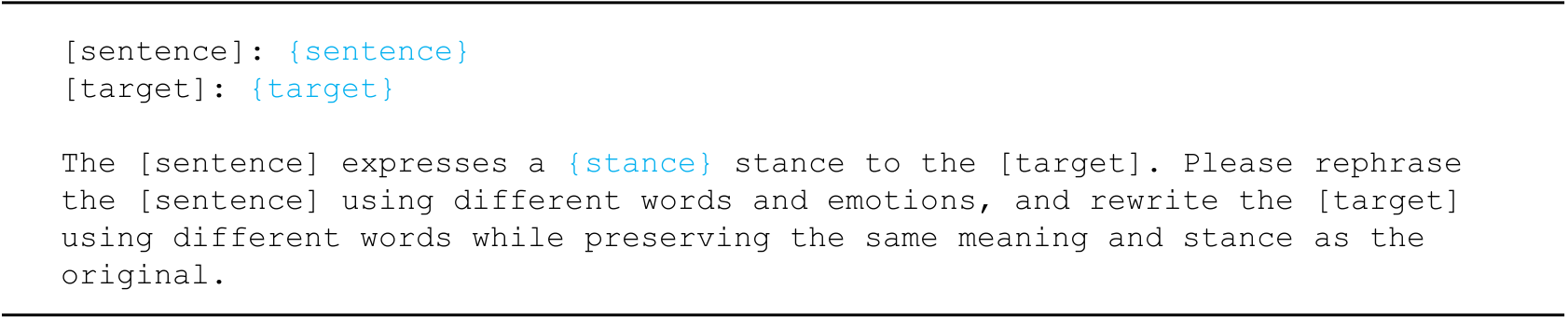}
\caption{Prompt template that allows the LLMs to rephrase the original sentence with different words and sentiments and express the target while ensuring that the semantics and the stance of the perturbed sample towards the target remain unchanged. Fill the blue text with the corresponding text, target, and stance label from the sample.}
\label{fig-prompt-non-causal}
\end{figure*}

\begin{figure*}[t]
\centering  
\includegraphics[width=1\linewidth]{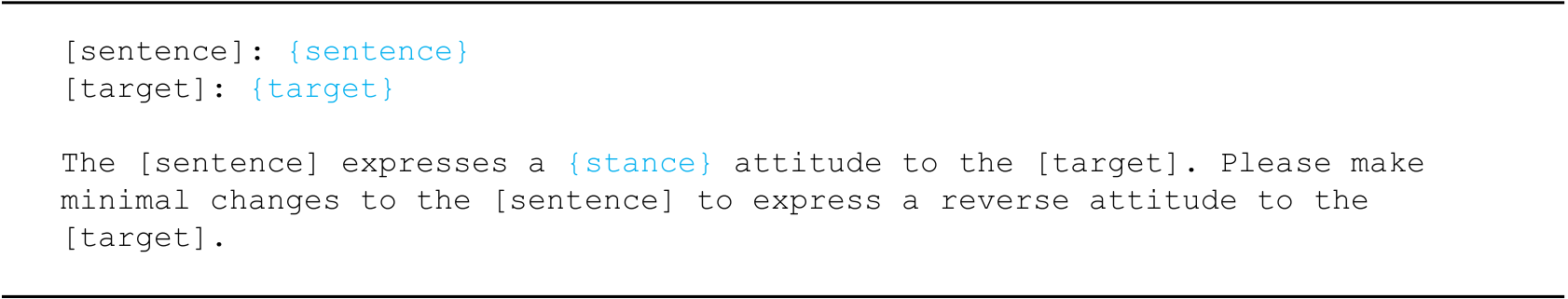}
\caption{Prompt template that makes necessary modifications to reverse the applicability of the label. Fill the blue text with the corresponding text, target, and stance label from the sample.}
\label{fig-prompt-causal}
\end{figure*}

Figures~\ref{fig-prompt-non-causal} and~\ref{fig-prompt-causal} illustrate the prompt templates used to obtain non-causal and causal counterfactual augmented data, respectively, as discussed in Section~\ref{subsec:cda}. Figure~\ref{fig-prompt-non-causal} presents the constructed instruction for acquiring non-causal counterfactual augmented data, while Figure~\ref{fig-prompt-causal} shows the instruction for obtaining causal counterfactual augmented data.

\section{Experimental Result of LLMs Bias}
\label{sec:experimental_result}

\begin{table}[t]
\small
\centering
\setlength{\tabcolsep}{5pt}
\renewcommand{\arraystretch}{1.1}
\begin{tabular}{lccc}
\hline
                   & \multicolumn{3}{c}{P-Stance}  \\
                   \cline{2-4}
                   & Avg$\uparrow$         & SSC$\downarrow$      & TPB$\downarrow$ \\
\hline
\rowcolor[gray]{0.92} FACTUAL$_{\text{LLaMA-2}}$     & \textbf{73.82}$^\star$ & \textbf{9.61}  & \textbf{5.52}$^\star$ \\
w/o Calibration    & 67.96          & 27.52          & 27.56  \\
\cdashline{1-4}[2pt/3pt]
w/o CAD            & 71.61          & 15.43          & 12.07  \\
- w/o non-causal & 71.29          & 13.12          & 13.25    \\
- w/o causal     & 69.39          & 10.66          & 14.82    \\
\hline
\rowcolor[gray]{0.92} FACTUAL$_{\text{GPT-3.5}}$     & \textbf{76.16}$^\star$ & 11.31          & \textbf{7.74}$^\star$ \\
w/o Calibration    & 73.32          & 18.08          & 13.47  \\
\cdashline{1-4}[2pt/3pt]
w/o CAD            & 70.39          & 12.92          & 11.83  \\
- w/o non-causal & 74.08          & 14.36          & 11.80    \\
- w/o causal     & 71.71          & \textbf{10.22} & 10.21    \\
\hline
\end{tabular}
\caption{Experimental results of ablation study of in-target stance detection on the P-Stance dataset. The best scores are in bold. Results with $\star$ indicate the significance tests of our \texttt{FACTUAL} over the ablation experiments at p-value < 0.05.}
\label{Tab:ablation-results-p-stance}
\end{table}

Table~\ref{Tab:ablation-results-p-stance} shows the experimental results of the ablation study on the P-Stance dataset, which are consistent with our previous observations.
We present the complete experimental results in Section~\ref{bias-llms}. Tables~\ref{Tab:sem16-sentiment-bias},~\ref{Tab:pstance-sentiment-bias}, and~\ref{Tab:vast-sentiment-bias} show the RStd and the macro F1-Score of samples with different sentiment on the Sem16, P-Stance, and VAST datasets. Tables~\ref{Tab:sem16-target-bias} and~\ref{Tab:pstance-vast-target-bias} present the RStd and the macro F1-Score of samples with different stance target on the Sem16, P-Stance, and VAST datasets. We can observe that in most cases, a larger stance bias leads to poorer stance detection results. In Tables~\ref{Tab:sem16-sentiment-bias},~\ref{Tab:pstance-sentiment-bias}, and~\ref{Tab:vast-sentiment-bias}, samples with positive and negative emotions exhibited larger Rstd, indicating that sentiment influenced the stance judgment of LLMs as a bias pattern. In Tables ~\ref{Tab:sem16-target-bias} and~\ref{Tab:pstance-vast-target-bias}, on some controversial debate topics such as the "Legalization of Abortion", the "Feminist Movement", and specific individuals like "Donald Trump" and "Hillary Clinton", larger Rstd indicates that LLMs demonstrated a relatively large target preference bias.

\begin{table*}[t]
\small
\centering
\setlength{\tabcolsep}{2.0pt}
\renewcommand{\arraystretch}{1.1}
\begin{tabular}{lcccccccc}
\hline
    & \multicolumn{8}{c}{Sem16(\%)}                                
\\ 
\cline{2-9} 
    & \multicolumn{2}{c}{Positive} &    & \multicolumn{2}{c}{Neutral} &    & \multicolumn{2}{c}{Negative}
\\
\cline{2-3} \cline{5-6} \cline{8-9}
     & RStd$\downarrow$ & F1$\uparrow$ &  & RStd$\downarrow$ & F1$\uparrow$ &   & RStd$\downarrow$ & F1$\uparrow$ 
\\
\hline
\multicolumn{9}{l}{\textbf{LLaMA-2-70b-chat}}                                                               
\\
Task-Des  & \textbf{22.20} & \textbf{59.73} &  & \textbf{23.65} & 58.11 &  & 7.56  & 65.19
\\
CoT-Demo  & 26.15 & 58.24 &  & 27.88 & 55.18 &  & 28.54 & 62.04
\\
Debias-Instruct & 26.86 & 55.04 &  & 25.51 & \textbf{58.20} &  & \textbf{5.34}  & \textbf{68.61}
\\
\hline
\multicolumn{9}{l}{\textbf{GPT-3.5-Turbo-0125}}                                                               
\\
Task-Des & 29.37 & 48.37 &  & 25.22 & 58.78 &  & 26.80  & 55.44
\\
CoT-Demo & \textbf{16.02} & \textbf{65.77} &  & 25.47 & \textbf{61.69} &  & \textbf{12.74} & \textbf{70.69}
\\
Debias-Instruct & 30.52 & 46.09 &  & \textbf{16.31} & 60.07  &  & 24.43 & 54.91
\\
\hline

\end{tabular}
\caption{RStd of sentiment labels and macro F1-score of stance detection on Sem16 dataset. The best scores are in bold.}
\label{Tab:sem16-sentiment-bias}
\end{table*}

\begin{table*}[t]
\small
\centering
\setlength{\tabcolsep}{2.0pt}
\renewcommand{\arraystretch}{1.1}
\begin{tabular}{lcccccccc}
\hline
    & \multicolumn{8}{c}{P-Stance(\%)}                                
\\ 
\cline{2-9} 
    & \multicolumn{2}{c}{Positive} &    & \multicolumn{2}{c}{Neutral} &    & \multicolumn{2}{c}{Negative}
\\
\cline{2-3} \cline{5-6} \cline{8-9}
     & RStd$\downarrow$ & F1$\uparrow$ &  & RStd$\downarrow$ & F1$\uparrow$ &   & RStd$\downarrow$ & F1$\uparrow$ 
\\
\hline
\multicolumn{9}{l}{\textbf{LLaMA-2-70b-chat}}                                                               
\\
Task-Des & 33.32 & \textbf{68.42} &  & 19.98 & 62.68 &  & \textbf{16.77} & 72.97
\\
CoT-Demo & \textbf{32.62} & 67.17 &  & \textbf{15.10} & \textbf{65.00} &  & 20.70 & \textbf{73.61}
\\
Debias-Instruct & 34.66 & 66.49 &  & 22.81 & 59.14 &  & 17.10 & 71.82
\\
\hline
\multicolumn{9}{l}{\textbf{GPT-3.5-Turbo-0125}}                                                               
\\
Task-Des & 33.79 & \textbf{68.61} &  & 23.48 & 62.32 &  & \textbf{13.88} & \textbf{75.66}
\\
CoT-Demo & \textbf{32.39} & 66.99 &  & \textbf{19.25} & \textbf{65.09} &  & 16.60 & 74.44
\\
Debias-Instruct & 33.56 & 68.00 &  & 20.05 & 63.19 &  & 16.83 & 75.07
\\
\hline

\end{tabular}
\caption{RStd of sentiment labels and macro F1-score of stance detection on P-Stance dataset. The best scores are in bold.}
\label{Tab:pstance-sentiment-bias}
\end{table*}

\begin{table*}[t]
\small
\centering
\setlength{\tabcolsep}{2.0pt}
\renewcommand{\arraystretch}{1.1}
\begin{tabular}{lcccccccc}
\hline
    & \multicolumn{8}{c}{VAST(\%)}                                
\\ 
\cline{2-9} 
    & \multicolumn{2}{c}{Positive} &    & \multicolumn{2}{c}{Neutral} &    & \multicolumn{2}{c}{Negative}
\\
\cline{2-3} \cline{5-6} \cline{8-9}
     & RStd$\downarrow$ & F1$\uparrow$ &  & RStd$\downarrow$ & F1$\uparrow$ &   & RStd$\downarrow$ & F1$\uparrow$ 
\\
\hline
\multicolumn{9}{l}{\textbf{LLaMA-2-70b-chat}}                                                               
\\
Task-Des & \textbf{30.19} & \textbf{61.47} &  & \textbf{5.79}  & 60.85 &  & 14.63 & 66.97
\\
CoT-Demo & 36.20 & 59.60 &  & 18.93 & \textbf{67.87} &  & \textbf{12.51} & 63.87
\\
Debias-Instruct & 34.60 & 56.63 &  & 10.82 & 59.49 &  & 13.48 & \textbf{67.61}
\\
\hline
\multicolumn{9}{l}{\textbf{GPT-3.5-Turbo-0125}}                                                               
\\
Task-Des & 36.33 & 44.58 &  & 19.75 & 52.72 &  & 30.03 & 46.74
\\
CoT-Demo & \textbf{26.01} & \textbf{69.05} &  & \textbf{8.78}  & \textbf{66.73} &  & \textbf{14.17} & \textbf{67.38}
\\
Debias-Instruct & 38.40 & 40.96 &  & 23.51 & 49.01 &  & 29.66 & 46.19
\\
\hline

\end{tabular}
\caption{RStd of sentiment labels and macro F1-score of stance detection on VAST dataset. The best scores are in bold.}
\label{Tab:vast-sentiment-bias}
\end{table*}

\begin{table*}[t]
\small
\centering
\setlength{\tabcolsep}{2.0pt}
\renewcommand{\arraystretch}{1.1}
\begin{tabular}{lcccccccccccccc}
\hline
    & \multicolumn{14}{c}{Sem16(\%)}                                
\\ 
\cline{2-15} 
    & \multicolumn{2}{c}{HC} &    & \multicolumn{2}{c}{FM} &    & \multicolumn{2}{c}{LA} &    & \multicolumn{2}{c}{A} &   & \multicolumn{2}{c}{CC}
\\
\cline{2-3} \cline{5-6} \cline{8-9} \cline{11-12} \cline{14-15}
     & RStd$\downarrow$ & F1$\uparrow$ &  & RStd$\downarrow$ & F1$\uparrow$ &  & RStd$\downarrow$ & F1$\uparrow$ &  & RStd$\downarrow$ & F1$\uparrow$ &  & RStd$\downarrow$ & F1$\uparrow$
\\
\hline
\multicolumn{15}{l}{\textbf{LLaMA-2-70b-chat}}                                                               
\\
Task-Des        & 11.59 & 73.64 &  & 24.41 & 58.33 &  & \textbf{6.84}  & 59.36 &  & \textbf{15.12} & 48.94 &  & 29.98 & 60.15 \\
CoT-Demo        & 28.36 & 64.71 &  & 35.98 & 52.96 &  & 19.27 & 57.10 &  & 19.91 & \textbf{55.87} &  & 34.26 & 62.78 \\
Debias-Instruct & \textbf{8.73}  & \textbf{76.50} &  & \textbf{14.75} & \textbf{63.39} &  & 9.77  & \textbf{58.64} &  & 25.76 & 39.32 &  & \textbf{22.83} & \textbf{69.16} \\
\hline
\multicolumn{15}{l}{\textbf{GPT-3.5-Turbo-0125}}                                                               
\\
Task-Des        & 28.68 & 61.97 &  & 24.21 & 57.47 &  & 26.98 & 56.75 &  & 5.81  & 27.54 &  & 27.52 & 60.36 \\
CoT-Demo        & \textbf{20.99} & \textbf{74.12} &  & \textbf{15.16} & \textbf{65.61} &  & \textbf{11.12} & \textbf{64.74} &  & 6.75  & \textbf{57.89} &  & \textbf{13.30} & \textbf{75.57} \\
Debias-Instruct & 28.07 & 63.07 &  & 27.81 & 54.72 &  & 24.08 & 57.11 &  & \textbf{5.43}  & 29.14 &  & 23.94 & 62.63 \\
\hline

\end{tabular}
\caption{RStd of targets and macro F1-score of stance detection on Sem16 dataset. The best scores are in bold.}
\label{Tab:sem16-target-bias}
\end{table*}

\begin{table*}[t]
\small
\centering
\setlength{\tabcolsep}{2.0pt}
\renewcommand{\arraystretch}{1.1}
\begin{tabular}{lccccccccccc}
\hline
    & \multicolumn{8}{c}{P-Stance(\%)} &    & \multicolumn{2}{c}{VAST(\%)}   
\\ 
\cline{2-10} \cline{11-12} 
    & \multicolumn{2}{c}{JB} &    & \multicolumn{2}{c}{BS} &    & \multicolumn{2}{c}{DT} &    & \multicolumn{2}{c}{ALL}
\\
\cline{2-3} \cline{5-6} \cline{8-9} \cline{11-12}
     & RStd$\downarrow$ & F1$\uparrow$ &  & RStd$\downarrow$ & F1$\uparrow$ &  & RStd$\downarrow$ & F1$\uparrow$ &  & RStd$\downarrow$ & F1$\uparrow$
\\
\hline
\multicolumn{12}{l}{\textbf{LLaMA-2-70b-chat}}                                                               
\\
Task-Des        & 2.69 & 84.31 &  & 8.79 & 77.29       &  & \textbf{15.79} & 78.08 &  & 7.76  & 68.36 \\
CoT-Demo        & 7.91 & \textbf{85.03} &  & \textbf{7.62} & \textbf{79.77}       &  & 19.19 & 77.52 &  & 9.64  & 67.08 \\
Debias-Instruct & \textbf{1.58} & 82.63 &  & 9.29 & 75.10       &  & 15.96 & \textbf{78.36} &  & \textbf{4.86}  & \textbf{69.10} \\
\hline
\multicolumn{12}{l}{\textbf{GPT-3.5-Turbo-0125}}                                                               
\\
Task-Des        & \textbf{0.53} & \textbf{83.20} &  & 4.82 & \textbf{80.02} &  & \textbf{10.94} & 81.66 &  & 28.44 & 49.86 \\
CoT-Demo        & 3.22 & 83.07 &  & \textbf{4.01} & 77.98       &  & 12.59 & 81.59 &  & \textbf{8.40}  & \textbf{69.90} \\
Debias-Instruct & 0.63 & 82.91 &  & 5.36 & 79.01       &  & 11.39 & \textbf{82.85} &  & 26.77 & 51.66 \\
\hline

\end{tabular}
\caption{RStd of targets and macro F1-score of stance detection on P-Stance and VAST dataset. The best scores are in bold.}
\label{Tab:pstance-vast-target-bias}
\end{table*}

\section{Prompt Robustness}
\label{sec:prompt_robustness}
We conducted experiments, and demonstrating that our method is robust across different prompt templates. The results are shown in Table~\ref{Tab:prompt_robustness}, and the prompts used are listed in Table~\ref{tab:three-prompts}.
According to experimental results, the variance in direct stance inference through prompts (w/o Calibration) is 3.5454 for LLaMA-2 and 10.5929 for GPT-3.5-Turbo. In contrast, the variance for stance judgments output by our \texttt{FACTUAL} is 0.3766 for LLaMA-2 and 1.1055 for GPT-3.5-Turbo. We believe this robustness stems from the fact that our calibration network uses LLM-generated rationales to analyze the stance of samples.

\begin{table*}[t]
\small
\centering
\setlength{\tabcolsep}{6pt}
\renewcommand{\arraystretch}{1.1}
\begin{tabular}{lcccccc}
\hline
\multicolumn{1}{c}{} & \multicolumn{3}{c}{Sem16 (LLaMA-2-70b-chat)}   & \multicolumn{3}{c}{Sem16 (GPT-3.5-Turbo-0125)}  \\
\cline{2-7}
                     & Avg            & Bias-SSC      & Bias-TPB      & Avg            & Bias-SSC       & Bias-TPB      \\
\hline
\rowcolor[gray]{0.92} FACTUAL-Prompt-1      & \textbf{73.82} & \textbf{9.61} & \textbf{5.52} & \textbf{76.16} & 11.31          & \textbf{7.74} \\
w/o Calibrition    & 67.96          & 27.52         & 27.56         & 73.32          & 18.08          & 13.47         \\
w/o CAD            & 71.61          & 15.43         & 12.07         & 70.39          & 12.92          & 11.83         \\
\cdashline{1-7}[2pt/3pt]
- w/o CAD-non-causal & 71.29          & 13.12         & 13.25         & 74.08          & 14.36          & 11.80         \\
- w/o CAD-causal     & 69.39          & 10.66         & 14.82         & 71.71          & \textbf{10.22} & 10.21         \\
\hline
\rowcolor[gray]{0.92} FACTUAL-Prompt-2      & \textbf{72.48} & \textbf{9.99} & \textbf{7.69} & \textbf{74.73} & \textbf{10.86} & \textbf{8.77} \\
w/o Calibrition    & 66.97          & 26.92         & 23.75         & 69.50          & 22.22          & 15.88         \\
w/o CAD            & 70.51          & 19.26         & 13.58         & 70.49          & 19.89          & 12.40         \\
\cdashline{1-7}[2pt/3pt]
- w/o CAD-non-causal & 71.00          & 11.27         & 13.27         & 72.01          & 15.51          & 10.05         \\
- w/o CAD-causal     & 69.89          & 10.03         & 11.57         & 70.81          & 11.58          & 10.56         \\
\hline
\rowcolor[gray]{0.92} FACTUAL-Prompt-3      & \textbf{73.74} & 9.19          & \textbf{5.60} & \textbf{73.59} & 10.11          & \textbf{7.77} \\
w/o Calibrition    & 67.37          & 19.45         & 18.14         & 65.35          & 26.35          & 25.88         \\
w/o CAD            & 69.99          & 12.83         & 13.15         & 67.86          & 16.84          & 18.24         \\
\cdashline{1-7}[2pt/3pt]
- w/o CAD-non-causal & 72.65          & 9.47          & 15.46         & 71.51          & 13.70          & 16.03         \\
- w/o CAD-causal     & 69.90          & \textbf{8.11} & 10.27         & 69.60          & \textbf{9.53}  & 13.16         \\
\hline
\end{tabular}
\caption{Experimental results of three different prompt templates on in-target stance detection on the Sem16 dataset. The best scores are in bold.}
\label{Tab:prompt_robustness}
\end{table*}

\begin{table*}[t]
\small
\centering
\setlength{\tabcolsep}{2.5pt}
\renewcommand{\arraystretch}{1.1}
\begin{tabular}{c|m{13cm}}
\hline
Task Description & Stance detection is to determine the attitude or tendency towards a certain target through a given sentence, including favor, against, and neutral. **Please read the following examples carefully and use them as references to judge the attitude of the sentence towards the target.** \\
\hline
Prompt \#1 & [Task Description]\textbackslash n \textbackslash n[in-context examples]\textbackslash n Your sentence: {sentence}\textbackslash n Question: What is the attitude of the sentence toward "{target}"? Please select the correct answer from "favor", "against" and "neutral". Answer this question with JSON format: ```json { "answer": "your answer", "stance": "favor" | "against" | "neutral" }``` \\
\hline
Prompt \#2 & [Task Description]\textbackslash n\textbackslash n[in-context examples]\textbackslash n Give you [sentence] and [target], please judge the attitude of [sentence] toward [target]. Select the correct answer from favor, against and neutral.\textbackslash n [sentence]: {sentence}\textbackslash n [target]: {target}\textbackslash n Only select the correct answer from "favor", "against" and "neutral".\textbackslash n Answer this question with JSON format: ```json { "answer": "your answer", "stance": "favor" | "against" | "neutral" }``` \\
\hline
Prompt \#3 & [Task Description]\textbackslash n\textbackslash n[in-context examples]\textbackslash n What is the attitude of [sentence] toward [target]?\textbackslash n A.against\textbackslash n B.favor\textbackslash n C.neutral\textbackslash n[sentence]: {sentence}\textbackslash n[target]: {target}\textbackslash n Answer this question with JSON format: ```json { "answer": "your answer", "stance": "A.against" | "B.favor" | "C.neutral" } ``` \\
\hline
\end{tabular}
\caption{Three different prompts used in the experiment in Table~\ref{Tab:prompt_robustness}.}
\label{tab:three-prompts}
\end{table*}

\section{Human Evaluation of Counterfactual Augmented Data}
\label{sec:human_eval}

\begin{table*}[t]
\small
\centering
\setlength{\tabcolsep}{2.5pt}
\renewcommand{\arraystretch}{1.1}
\begin{tabular}{lccccc}
\hline
           & \multicolumn{2}{c}{CAD-non-causal} & & \multicolumn{2}{c}{CAD-causal} \\
           \cline{2-3} \cline{5-6}
           & Quality        & Achievement       & & Quality      & Achievement     \\
\hline
Human Eval & 96.67\%          & 91.33\%             & & 95.33\%        & 88.83\% \\
\hline
\end{tabular}
\caption{Human evaluation on Sem16, P-Stance, and VAST datasets. Quality stands for qualitative assessment of the generated samples and Achievement stands for the achievement of generating objectives.}
\label{Tab:human-eval}
\end{table*}

We randomly select 500 samples and use human evaluation (with three experienced researchers who are not involved in this work and have worked on natural language processing for over 3 years) to measure the counterfactual data generated by GPT-3.5-turbo. 
The primary consideration focuses on the qualitative assessment of the generated samples, necessitating evaluators to confirm the accuracy of both the grammar and the affirmed stance.
The secondary consideration pertains to achieving generating objectives, necessitating evaluators to confirm if the samples were generated as guideline instructions. Evaluators respond to these considerations with a binary "yes" or "no". Subsequently, we calculate the average ratio of affirmative responses from three evaluators for each query. The results in Table~\ref{Tab:human-eval} show that the generated samples are of high quality, contributing substantially to our calibration network training.

\section{Case Study}
\label{sec:case_study}
We conduct a case study on Sem16, P-Stance and VAST datasets, to analyze the biases of LLMs in the stance detection task and the practical effectiveness of our calibration network. The results are show in Table~\ref{tab:case-sem16}, \ref{tab:case-pstance} and \ref{tab:case-vast}.
The correct analysis patterns of LLMs are marked in blue, while biased analysis patterns are marked in red. 
We can observe that for some samples with strong sentiment expressions, such as the examples in Table~\ref{tab:case-pstance}, LLMs are influenced by sentiment Spurious cues and result in biased stance judgments. For some controversial debate topics, such as the examples in Table~\ref{tab:case-sem16}, LLMs generate hallucinations due to their preferences, leading to biased stance judgments.

\begin{table*}[t]
\small
\centering
\setlength{\tabcolsep}{2.5pt}
\renewcommand{\arraystretch}{1.1}
\begin{tabular}{m{5cm}|c|c|m{6cm}|c}
\hline
\multicolumn{1}{c|}{Text (Target: Legalization of Abortion)} & Label & LLMs Pred & \multicolumn{1}{c|}{LLMs Stance Rationale} & Calibration \\
\hline
A prochoice advocate but circumcise ur baby? Fucking hypocrite! \#circumcision \#humanrights & Against & Favor & The sentence criticizes the hypocrisy of being a prochoice advocate but supporting circumcision, indicating a \textcolor{blue}{negative attitude towards circumcising babies.} \textcolor{red}{Since the author is prochoice, it can be inferred that they are in favor of the legalization of abortion.} & Against \\
\hline
\multicolumn{1}{c|}{Text (Target: Feminist Movement)} & Label & LLMs Pred & \multicolumn{1}{c|}{LLMs Stance Rationale} & Calibration \\
\hline
Equality is the police burying a domestic violence accusation against a female sports star, too \#wedidit \#usa  & Against & Favor & The sentence mentions burying a domestic violence accusation against a female sports star, which implies a negative view towards this unequal treatment. \textcolor{red}{Therefore, the attitude of the sentence towards the Feminist Movement is likely favorable.} & Against \\
\hline
\end{tabular}
\caption{Case study on the Sem16 dataset.}
\label{tab:case-sem16}
\end{table*}

\begin{table*}[t]
\small
\centering
\setlength{\tabcolsep}{2.5pt}
\renewcommand{\arraystretch}{1.1}
\begin{tabular}{m{5cm}|c|c|m{6cm}|c}
\hline
\multicolumn{1}{c|}{Text (Target: Donald Trump)} & Label & LLMs Pred & \multicolumn{1}{c|}{LLMs Stance Rationale} & Calibration \\
\hline
You might as well put a sock in it. Voters see these coordinated stupid attacks are like your \#Kavanaugh symphony. \#Dems can't win with hollow attempts to bash \#Trump. & Favor & Against & The sentence uses derogatory terms like 'stupid attacks' and 'hollow attempts to bash \#Trump,' indicating a critical perspective.  Therefore, \textcolor{red}{the attitude of the sentence toward 'Donald Trump' is against.} & Favor \\
\hline
\multicolumn{1}{c|}{Text (Target: Joe Biden)} & Label & LLMs Pred & \multicolumn{1}{c|}{LLMs Stance Rationale} & Calibration \\
\hline
Saying that is as absurd as saying breast cancer screening increases breast cancer. Is there any Republican who isn't as dumb as dirt?  & Favor & Against & The sentence \textcolor{blue}{implies negative assumptions about the intelligence of Republicans}. \textcolor{red}{The mention of Joe Biden in the context of criticizing Republicans suggests a critical or skeptical attitude towards Joe Biden.} & Favor \\
\hline
\end{tabular}
\caption{Case study on the P-Stance dataset.}
\label{tab:case-pstance}
\end{table*}

\begin{table*}[t]
\small
\centering
\setlength{\tabcolsep}{2.5pt}
\renewcommand{\arraystretch}{1.1}
\begin{tabular}{m{5cm}|c|c|m{6cm}|c}
\hline
\multicolumn{1}{c|}{Text (Target: Homosexual)} & Label & LLMs Pred & \multicolumn{1}{c|}{LLMs Stance Rationale} & Calibration \\
\hline
Yet that same Bible gives us Christians a particular anthropology, and reveals to us the meaning of sex and marriage within the cosmic order. And herein lies the problem. If you look to a "holy" book written long, long ago and betraying its human origins on every page for guidance on issues of cosmology, biology, sexuality and other human and natural sciences, you will inevitably be in constant conflict with social progress and, indeed, reality. There is no secular opposition to homosexuality." & Favor & Against & The sentence \textcolor{blue}{criticizing the use of a 'holy' book for guidance on issues related to sexuality and natural sciences.} It suggests that \textcolor{blue}{relying on such sources will lead to conflict with social progress and reality.} Additionally, \textcolor{blue}{the statement asserts that there is no secular opposition to homosexuality, implying a negative view towards those who oppose it based on religious beliefs}. Overall, \textcolor{red}{the tone of the sentence is against homosexuality, indicating a critical stance.} & Favor \\
\hline
\multicolumn{1}{c|}{Text (Target: Gun Control)} & Label & LLMs Pred & \multicolumn{1}{c|}{LLMs Stance Rationale} & Calibration \\
\hline
Military and police, yes. They're sworn and trained to uphold the country and the laws of the country. The rest is an ego trip.  & Favor & Against & The sentence expresses a negative attitude towards gun control in the military and police, stating that they are sworn and trained to uphold the country and its laws. However, \textcolor{blue}{it is critical of the use of guns by others, referring to it as an ego trip.} \textcolor{red}{Therefore, the attitude of the sentence towards gun control is against} & Favor \\
\hline
\end{tabular}
\caption{Case study on the VAST dataset.}
\label{tab:case-vast}
\end{table*}

\end{document}